\documentclass{article}

\usepackage[table]{xcolor}
\usepackage[preprint]{corl_2026} 
\usepackage{amsmath}
\usepackage{amssymb}
\usepackage{graphicx}
\usepackage{caption}
\usepackage{wrapfig}
\usepackage{enumitem}
\usepackage{booktabs}
\usepackage{multirow}
\usepackage{pifont}
\newcommand{\cmark}{\ding{51}}
\newcommand{\xmark}{\ding{55}}
\definecolor{mydeepred}{RGB}{150,40,40}
\title{W2A: Distilling World Priors into Action Models for Real-Time Robotic Control}

\author{
  Xiaopeng~Zhang, Yongjin Mu, Yueyang Weng, Chengzhi Zhao, Yanjie Li\textsuperscript{\ensuremath{\ddagger}} \\ \\
  Harbin Institute of Technology, Shenzhen \\ \\
  \textsuperscript{\ensuremath{\ddagger}} Corresponding author \\
}

\begin{document}
\maketitle


\begin{figure*}[h]
    \centering
    \includegraphics[width=\textwidth]{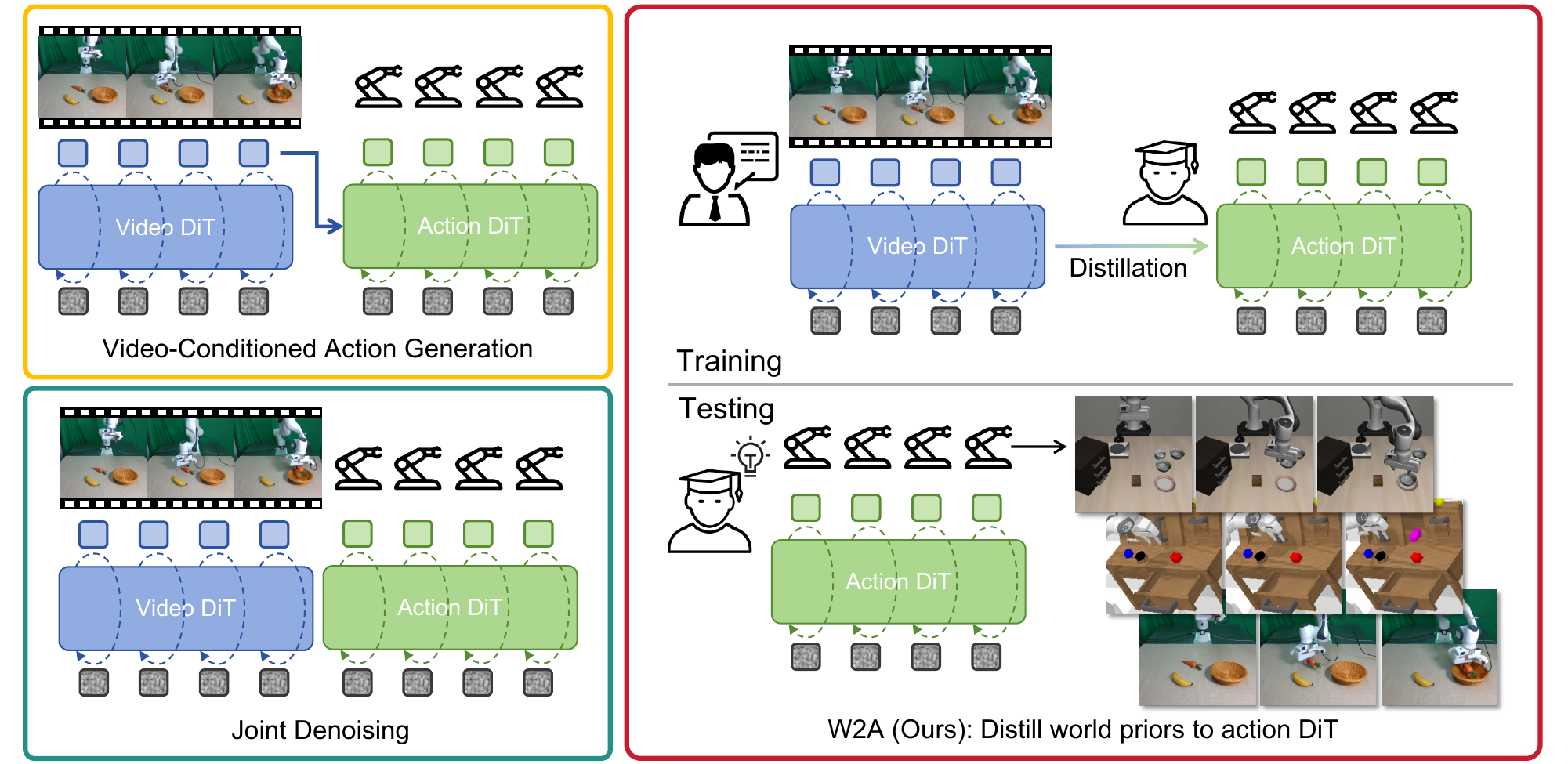} 
    \caption{Overview of W2A compared with mainstream world action models. While existing WAMs usually keep a video branch at test time, W2A distills world priors from a video DiT into an action DiT during training and removes the video DiT entirely at test time. This yields competitive performance in simulation and real-world manipulation while substantially improving inference efficiency.}
    \label{fig:overview}
\end{figure*}

\begin{abstract}
    World Action Models (WAMs) have recently shown great promise in robotic control, benefiting from the future imagination capability of large-scale pretrained video models. However, most existing WAMs require explicitly imagining future states at test time, incurring substantial computational overhead and latency that make them unsuitable for real-time robotic control. In this paper, we propose \textbf{W2A}, a two-stage distillation framework for real-time robotic manipulation that distills intermediate denoising representations from a video diffusion model into an action model during training, while entirely discarding the video model at test time. We hypothesize that although video latent representations are high-dimensional, a compressed low-dimensional subspace is sufficient to provide effective supervision for action generation. This makes it possible to transfer such information effectively into an action model. Extensive experiments in both simulation and real-world settings demonstrate competitive success rates while significantly improving inference efficiency. On LIBERO, W2A achieves a 95.9\% average success rate. More importantly, W2A achieves about 21$\times$ faster inference than the representative WAM baseline. Project website: \url{https://misaka0502.github.io/W2A}.
\end{abstract}



\section{Introduction}
\label{sec:introduction}
	Building general robotic policies is a central goal on the path toward artificial general intelligence. Built upon internet-scale pretrained Vision-Language Models (VLMs), Vision-Language-Action models (VLAs) have emerged as a promising paradigm for robotic control, offering strong semantic grounding and generalization~\citep{kim2024openvla,black2024pi_0,intelligence2025pi_,gr00tn1_2025,song2025reconvla}. However, because VLAs are largely trained on static image-text pretraining, they still have limited understanding of physical dynamics and procedures, both of which are crucial for long-horizon and contact-rich robotic manipulation.
    
    To address this limitation, recent works~\citep{hu2024video,pai2025mimicvideo,li2025unified,ma2026dit4dit,yuan2026fastwam} have begun to integrate video-based world modeling into robot policy learning, giving rise to World Action Models (WAMs). By leveraging future visual prediction or latent representations, WAMs incorporate richer temporal and physical structure. Existing WAMs can be broadly categorized into two classes: (1) video-conditioned action modeling and (2) joint modeling of future states and actions. The former injects predicted future observations or latent representations as conditions into an action expert for action generation~\citep{hu2024video,pai2025mimicvideo}. The latter, in contrast, jointly models future observations and actions during training, generating both within a unified framework~\citep{bi2025motusunifiedlatentaction,kim2026cosmos}. However, most existing WAMs require explicitly imagining future states at inference time, incurring substantial computational overhead and inference latency that make them unsuitable for real-time robotic control. 

    In this paper, we refer to \textbf{world priors} as the latent knowledge of video models about how scenes evolve over time. We focus on two questions: (1) what world priors in the video models used by WAMs are truly important for action generation, and whether they can be captured by a low-dimensional latent subspace? (2) Can such essential world priors be distilled into an action model, thereby eliminating the additional inference overhead and enabling real-time robotic control? Our key insight is that the latent representations of video models contain much information that is not directly useful for action generation, such as static scene content and task-irrelevant visual details, while only a small subset of features truly captures action-relevant dynamics. If this critical knowledge can be distilled into an action model, the video model can be removed at inference time.
    
    Motivated by this insight, we propose \textbf{W2A}, a two-stage distillation framework that transfers world priors from video models into action models for real-time robotic control. Specifically, in the first stage of training, we adapt a pretrained language-conditioned video diffusion transformer (DiT)~\citep{Peebles2022DiT} to robotic video data. We then apply principal component analysis (PCA)~\citep{hotelling1933analysis} to the latent representations of the video DiT, projecting them into a lower-dimensional latent space. In the second stage, we train an action DiT from scratch as our action model. For action DiT training, in addition to a flow matching objective~\citep{lipman2022flow}, we introduce a Q-Former~\citep{li2023blip} to read out features from the action DiT. These queried features are then aligned with the projected video latents through a similarity-based distillation loss. At test time, our framework removes the video DiT entirely, retaining only the action DiT. 
    We believe that although video latents and action latents are not directly causal variables, they share the same action-conditioned dynamics: one captures the resulting world evolution, while the other must encode the control patterns that induce it. This shared structure makes distilling world priors from video models into action models both practically and conceptually plausible. Figure~\ref{fig:overview} shows the differences between W2A and the existing WAMs.

    Our experimental results suggest that world priors useful for action generation can be effectively distilled into action models. This allows W2A to preserve the benefits of WAMs and improve the performance of action models, while avoiding the deployment overhead of video models.
    
\section{Related Work}
\label{sec:related_works}
    \textbf{Vision-Language-Action Models.} Vision-Language-Action models extend large VLMs from semantic understanding to embodied control, and have become a major direction for generalist robotic policies~\citep{zhao2025cot,shukor2025smolvla,team2025gemini,liu2024rdt,jiang2025galaxea,zhaozhang2025vla2,zhong2026acot,wu2026pragmatic,intelligence2026pi}. Early milestones such as RT-1~\citep{brohan2022rt} and RT-2~\citep{zitkovich2023rt} established the paradigm of mapping visual observations and language instructions to robot actions, while demonstrating that large-scale internet pretraining can significantly improve semantic grounding and task generalization in manipulation. Subsequent work shifted from task-specific instantiations toward more general and scalable policy learning. In particular, Octo~\citep{octo_2023} explored large-scale multi-task robot pretraining, and OpenVLA~\citep{kim2024openvla} further advanced this line by providing an open and reproducible VLA framework for general-purpose manipulation. More recent robot foundation models, including $\pi_{0.5}$~\citep{intelligence2025pi_}, $\pi^*_{0.6}$~\citep{intelligence2025pi}, and GR00T N1~\citep{gr00tn1_2025}, continue to push the frontier by scaling data diversity, improving cross-embodiment transfer, and strengthening real-world adaptability~\citep{team2025gemini}. 
    Despite these advances, VLAs are still largely rooted in static image-text pretraining, which limits their ability to reason about how actions change the world over time.

    \textbf{World Action Models and Video Generation for Policy Learning.} Recent progress in video generation has motivated new training paradigms that bring pretrained video world models into robotic control~\citep{yuan2026fastwam,du2023learning,wu2023unleashing,feng2025vidarembodiedvideodiffusion,zheng2026video,zhu2025uwm,team2026motubrain,guo2026unified,zhang2026world}. An early representative work is Video Prediction Policy (VPP)~\citep{hu2024video}, which showed that video diffusion models pretrained for future prediction can provide predictive visual representations that substantially improve downstream policy learning. Mimic-video~\citep{pai2025mimicvideo} retains the two-stage recipe of first adapting a video model and then training an action decoder, further demonstrating that video models can serve as strong spatio-temporal backbones for robot action generation. A different line pushes this integration further by reducing or removing the gap between world modeling and policy learning. Cosmos Policy~\citep{kim2026cosmos} directly adapts a pretrained video model into a robot policy with minimal architectural changes, jointly modeling actions, future states, and values within the latent diffusion process. More recent works such as LingBot-VA~\citep{lingbot-va2026}, DreamZero~\citep{ye2026worldactionmodelszeroshot}, and GigaWorld-Policy~\citep{ye2026gigaworld} further strengthen this trend by unifying future-state prediction and action generation within autoregressive or joint-denoising formulations. 
    However, most WAMs still retain a video backbone or world-modeling branch at test time, introducing additional computational overhead and inference latency for real-time control. W2A addresses this issue by using the video model only as a training-time teacher, distilling its world priors into an action model and discarding the video model entirely during deployment.

\section{Preliminaries}
\label{sec:preliminaries}
    \subsection{Problem Statement}
    \label{subsec:problem_formulation}
    We consider language-conditioned robotic manipulation from demonstrations. Let $o_t$ denote the current observation, $l$ the task description, and $a_{t:t+H-1}$ an action chunk of horizon $H$. A standard action policy directly predicts
    \begin{equation}
        a_{t:t+H-1} \sim \pi_\theta(\cdot \mid o_t, l).
    \end{equation}
    Many existing WAMs introduce an additional video model. Let $v$ denote imagined future visual states generated by the video model. Their action generation can be written as
    \begin{equation}
        p(a_{t:t+H-1}, v \mid o_t, l)
        =
        p_a(a_{t:t+H-1} \mid o_t, l, v)\,p_v(v \mid o_t, l) ,
    \end{equation}
    where $p_v$ and $p_a$ denote the probability distributions of video generation and action generation.
    
    In contrast, our formulation entirely removes the video model at test time. During training, we use a teacher video model to provide intermediate latent representations $Z^v$, 
    which serve as distillation targets for an action model. The deployed policy still follows the direct form $a_{t:t+H-1} \sim \pi_\theta(\cdot \mid o_t, l)$, while its hidden representations are shaped by video-based world priors during training.

    \begin{figure*}[t]
        \centering
        \includegraphics[width=\textwidth]{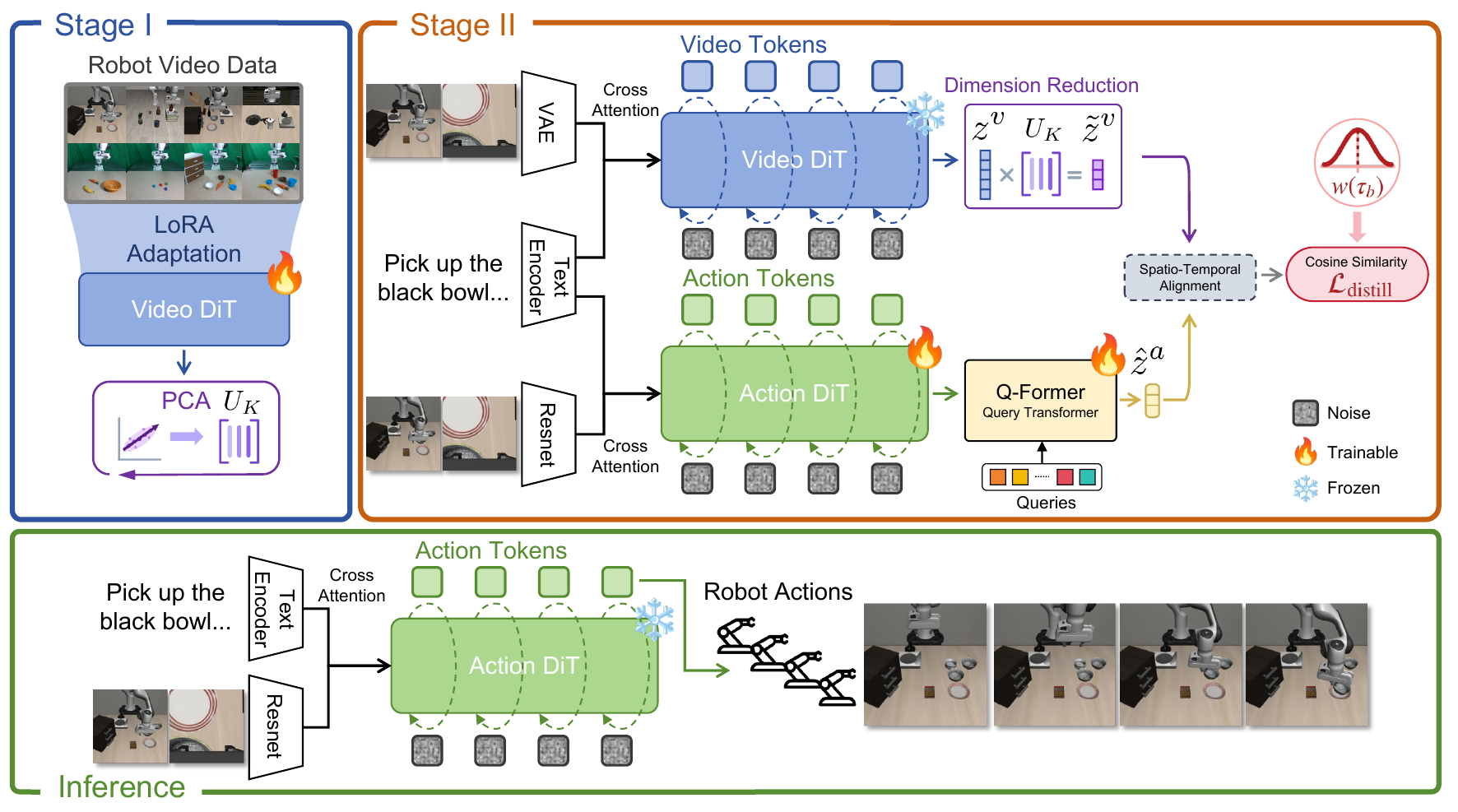} 
        \caption{The W2A training and inference framework.}
        \label{fig:framework}
    \end{figure*}

    \subsection{Flow Matching}
    \label{subsec:flow_matching}
    We adopt the Flow Matching framework~\citep{lipman2022flow} to train the models. Let $x_1 \sim \mathcal{N}(0,I)$ denote a Gaussian noise sample and let $x_0$ denote a target data sample. Following the flow matching parameterization, we define the interpolation path
    \begin{equation}
        x_\tau = \tau x_1 + (1-\tau)x_0, \qquad \tau \in [0,1].
    \end{equation}
    Under this convention, $\tau=1$ corresponds to pure noise and $\tau=0$ corresponds to the data sample.
    
    The action DiT is trained with a direct flow matching objective. Specifically, a conditional vector field $v_\theta(x_\tau,\tau)$ is learned to predict the target velocity
    \begin{equation}
        u(x_\tau,\tau) = x_1 - x_0,
    \end{equation}
    by minimizing
    \begin{equation}
        \mathcal{L}_{\mathrm{FM}}
        = \mathbb{E}_{x_0,x_1,\tau}
        \left[
            \left\| v_\theta(x_\tau,\tau) - (x_1-x_0) \right\|_2^2
        \right].
    \end{equation}
    At test time, samples are generated by integrating the learned vector field from $\tau=1$ to $\tau=0$:
    \begin{equation}
        \frac{d x_\tau}{d \tau} = v_\theta(x_\tau,\tau,c), \qquad x_1 \sim \mathcal{N}(0,I),
    \end{equation}
    which in practice can be implemented with a backward Euler update
    \begin{equation}
        x_{\tau_{k-1}} = x_{\tau_k} - \Delta \tau\, v_\theta(x_{\tau_k}, \tau_k),
    \end{equation}
    where $\Delta \tau = \tau_k - \tau_{k-1} > 0$. Moreover, the denoising time (or equivalently, the noise level) can be flexibly specified, a property that is directly exploited in our method to extract intermediate video latents at selected denoising stages as distillation targets.

\section{Method}
\label{sec:method}
        W2A is a two-stage distillation framework that distills world priors from a video DiT into an action DiT for real-time robotic control. In Stage I, we adapt a pretrained language-conditioned video DiT to robotic video data and compress its latent representations into a lower-dimensional space. In Stage II, we train an action DiT from scratch as our action model and augment its standard action modeling objective with a latent distillation objective, which aligns the action latents with the compressed video latents through a lightweight Q-Former bottleneck. At test time, the video DiT is entirely discarded, and only the action DiT is retained to generate actions conditioned on recent observations and task descriptions. The overall pipeline is illustrated in Figure~\ref{fig:framework}. 
    
    \subsection{Stage I: Video Domain Adaptation and PCA-Based Latent Compression}
        \label{subsec:stage_1}
        Stage I aims to obtain a compact video representation from a pretrained video DiT. We instantiate the video DiT as Cosmos-Predict2~\citep{nvidia2025cosmospredict2}, a large-scale language-conditioned video model, and adapt it to robotic video data via low-rank adaptation (LoRA)~\citep{hu2022lora}. This reduces the domain gap between generic internet videos and robot interaction data, making the resulting intermediate latents more aligned with manipulation-specific dynamics and task-relevant interaction patterns.
    
        Given the current observation $o_t$ and task description $l$, the adapted video DiT produces latent representations along its denoising trajectory. We extract video latents from an intermediate denoising stage and layer of the video DiT. Concretely, for each sample we extract an intermediate teacher video latent $Z^v \in \mathbb{R}^{T_v \times H_v \times W_v \times D_v}$, where $T_v$ denotes the number of temporal tokens, $H_v$ and $W_v$ denote the latent spatial resolution, and $D_v$ denotes the feature dimension. Let $z_{t,h,w}^v \in \mathbb{R}^{D_v}$ be the latent vector at temporal index $t$ and spatial location $(h,w)$. We then fit PCA over all $z_{t,h,w}^v$ and obtain the empirical mean $\mu \in \mathbb{R}^{D_v}$ together with the top-$K$ principal directions $U_K \in \mathbb{R}^{{D_v} \times K}$. Each teacher video latent vector is then projected into the low-dimensional subspace as
        \begin{equation}
            \tilde{z}_{t,h,w}^v = (z_{t,h,w}^v-\mu)U_K \in \mathbb{R}^{K}.
        \end{equation}
        Applying this projection to every spatio-temporal location yields the compressed $\tilde{Z}^v \in \mathbb{R}^{T_v \times H_v \times W_v \times K}$. This reduces only the feature dimension, while preserving its spatio-temporal structure. The resulting $\tilde{Z}^v$ provides compact video latents that are subsequently used as the distillation target in Stage II. The choices of the PCA dimension $K$ and the teacher latent extraction depth, together with their implementation settings, are discussed in Appendix~\ref{app:supplementary_ablation_experiments} and Appendix~\ref{app:model_structure}. To illustrate what information is preserved after PCA projection, we provide a visualization and discussion of raw and PCA-projected video latents in Appendix~\ref{app:visualization_and_discussion_of_raw_and_pca_projected_video_latents}.

    \subsection{Stage II: Action DiT Training and Latent Distillation with Q-Former}
    \label{subsec:stage_2}
        In Stage II, the action DiT is trained with a joint objective that combines the flow matching loss and a distillation loss from compact video latents. Similar to Stage I, we extract an action latent $Z^a \in \mathbb{R}^{H \times D_a}$ from an intermediate layer of the action DiT. Besides, we introduce a lightweight Q-Former~\citep{li2023blip} bottleneck with a set of learnable queries to extract from $Z^a$ the features for alignment with $\tilde{Z}^v$.  Formally, the queried student representation is
        \begin{equation}
            \hat{Z}^a = \mathcal{Q}_\psi(Q,Z^a),
        \end{equation}
        where $\psi$ denotes the Q-Former parameters and $Q$ denotes the learnable queries.
        
        To preserve the spatio-temporal structure of the teacher video latent, we organize the Q-Former outputs to match the layout of $\tilde{Z}^{v}$. Following the temporal compression ratio of the Cosmos-Predict2 VAE, we partition $Z^a$ into $G$ temporal groups, each corresponding to one video temporal token in $\tilde{Z}^{v}$. For the $g$-th group, a set of learnable spatial query tokens aggregates information from the action tokens in that group via cross-attention, producing an aligned action representation corresponding to the spatial grid of the compressed teacher video latent. Applying this operation across all groups yields $\hat{Z}^{a}$, whose spatio-temporal layout matches $\tilde{Z}^{v}$. Details are provided in Appendix~\ref{app:temporal_spatial_alignment}.
        
        
        We use a cosine distillation loss to align the queried action latent with the compressed video latent:
        \begin{equation}
            \mathcal{L}_{\mathrm{distill}}
            =
            \frac{1}{B}\sum_{b=1}^{B}
            w(\tau_b)\cdot
            \frac{1}{M}\sum_{i=1}^{M}
            \left(
            1-
            \frac{\langle \hat{z}^{a}_{b,i}, \tilde{z}^{v}_{b,i}\rangle}
            {\|\hat{z}^{a}_{b,i}\|_{2}\,\|\tilde{z}^{v}_{b,i}\|_{2}}
            \right),
        \end{equation}
        where $M = G \times H_v \times W_v$ is the total number of aligned spatio-temporal positions, $\tau_b \in [0,1]$ is the sampled flow time for the $b$-th training sample, and $\tau_c$ is the Gaussian center corresponding to the flow time at which the video latents are extracted. The weighting term
        \begin{equation}
            w(\tau)=\exp\!\left(-\frac{(\tau-\tau_c)^2}{2\sigma_w^2}\right)
        \end{equation}
        modulates the distillation strength over action flow time, assigning larger weights to action latents whose flow time is closer to $\tau_c$.
        
        The final training objective is
        \begin{equation}
            \mathcal{L}
            =
            \mathcal{L}_{\mathrm{FM}}
            +
            \lambda_{\mathrm{dist}} \mathcal{L}_{\mathrm{distill}},
        \end{equation}
        where $\mathcal{L}_{\mathrm{FM}}$ is the flow matching objective in Sec.~\ref{subsec:flow_matching}, and $\lambda_{\mathrm{dist}}$ controls the strength of distillation.

    \subsection{Inference for Real-Time Control}
    \label{subsec:inference}
        At test time, W2A retains only the action DiT, so action generation no longer depends on any explicit video generation. This yields a fully video-free deployment graph and eliminates the additional computational overhead and inference latency introduced by the video model.
    
        Given the current observation $o_t$ and task description $l$, the action DiT predicts a future action chunk of length $H$. Rather than executing the entire predicted sequence, the policy executes only the first $H_a$ actions, where $H_a < H$, and then replans based on the updated observations.
        
        To improve temporal continuity across replanning cycles, we adopt a warmstart mechanism~\citep{black2026real} that reuses the unexecuted suffix of the previous action chunk as a reference for the next prediction. Rather than sampling from pure Gaussian noise, this cached trajectory is perturbed to an intermediate flow time $\tau_{\mathrm{ws}}$ and refined under the latest observation and task condition. This enables W2A to maintain short-horizon consistency while adapting to new observations. 
        Overall, W2A achieves real-time control through video-free deployment, action chunking, and warmstarted replanning.

\section{Experiments}
\label{sec:experiments}
    We conduct comprehensive experiments in simulation and the real world to answer the following questions: (1) Does W2A improve policy performance? (2) What information is distilled from the video model into the policy? (3) How do key design choices contribute to the performance?

    \subsection{Experimental Setup}
    \label{subsec:experimental_setup}
        \textbf{Simulated Tasks.} We evaluate W2A on the simulated benchmarks LIBERO~\citep{liu2023libero} and CALVIN~\citep{mees2022calvin}, both of which are language-conditioned manipulation benchmarks. For LIBERO, we focus on four task suites: \textit{LIBERO-Spatial}, \textit{LIBERO-Object}, \textit{LIBERO-Goal}, and \textit{LIBERO-Long}, each containing 10 distinct tasks. For CALVIN, we focus on the challenging ABC$\rightarrow$D setting, where the policy is trained in environments A, B, and C, and evaluated in the unseen environment D. We follow the official evaluation protocol, evaluate with three random seeds, and report the average success rate.

        \textbf{Real-world Tasks.}
            We conduct real-world experiments using a Franka Research 3 robot equipped with a front-view RealSense D435i camera and a wrist-view RealSense D405 camera. We consider three task types with increasing difficulty: (1) pick-and-place (PnP), (2) cube-stacking (Cube), and (3) long-horizon (Long). Each task type contains four subtasks. In addition, we design three challenging tasks for more practical real-world scenarios: Towel folding, Table clearing, and USB insertion. For training, we collect 50 teleoperated demonstrations for each subtask using an Omega 7 device, resulting in 750 real-world demonstrations in total. During evaluation, each subtask is tested with 30 consecutive rollouts, and we report the average success rate for each task type. Figure~\ref{fig:task} shows several experimental scenarios.



    \subsection{Main Results}
    \label{subsec:main_results}
        \textbf{Simulated Tasks.}
        Table~\ref{tab:libero_main} and Table~\ref{tab:calvin_main} report the results on LIBERO and CALVIN. On LIBERO, W2A achieves 95.9\% average success rate and consistently outperforms other scratch-trained baselines. Moreover, it remains competitive with large finetuned foundation models, and even achieves the best result on LIBERO-Long (93.6\%), surpassing $\pi_{0.5}$. This suggests that distilled world priors are particularly helpful for long-horizon manipulation. On the CALVIN ABC$\rightarrow$D benchmark, W2A also demonstrates strong long-horizon instruction-following performance in unseen environments. Our method achieves an average task completion length of 4.30, remaining highly competitive with prior video-based policy learning methods. In particular, W2A attains the best success rates for completing the first two and all five consecutive subtasks, reaching 91.7\% and 77.1\%, respectively. Detailed per-seed results and standard deviation analysis are summarized in Appendix~\ref{app:detailed_simulation_results}.

        \begin{table*}[tp]
            \centering
            \caption{Results on LIBERO. ``finetuned'' denotes models trained on large-scale external robot action data and then finetuned on the target benchmark; ``scratch'' denotes models trained with robot action data only from the target evaluation benchmark, without extra robot action data. \textbf{Bold} indicates the best result overall, and \underline{underline} indicates the best result among ``scratch" models. Ext. RA denotes external robot-action data (\cmark: yes; \xmark: no).}
            \label{tab:libero_main}
            \renewcommand{\arraystretch}{1.15}
            \setlength{\tabcolsep}{7pt}
            \resizebox{1.0\linewidth}{!}{
            \begin{tabular}{lcccccc}
                \toprule
                Method & Ext. RA & Spatial (\%) & Object (\%) & Goal (\%) & Long (\%) & Average (\%) \\
                \midrule
                Diffusion Policy (scratch)~\citep{chi2023diffusionpolicy} & \xmark & 78.3 & 92.5 & 68.3 & 50.5 & 72.4 \\
                Octo (finetuned)                        & \cmark & 78.9 & 85.7 & 84.6 & 51.1 & 75.1 \\
                OpenVLA (finetuned)                & \cmark & 84.7 & 88.4 & 79.2 & 53.7 & 76.5 \\
                $\pi_{0.5}$ (finetuned)        & \cmark & \textbf{98.8} & 98.2 & \textbf{98.0} & 92.4 & \textbf{96.9} \\
                mimic-video (scratch)          & \xmark & 94.2 & 96.8 & 90.6 & 84.9 & 91.6 \\
                DreamZero-5B (scratch)~\citep{ye2026worldactionmodelszeroshot} & \xmark & 90.4 & 94.8 & 87.2 & 82.2 & 88.7 \\
                \midrule
                \rowcolor{gray!15}
                \textbf{W2A (ours, scratch)} & \xmark & \underline{96.3} & \underline{\textbf{98.8}} & \underline{95.0} & \underline{\textbf{93.6}} & \underline{95.9} \\
                \bottomrule
            \end{tabular}
            }
        \end{table*}

        \begin{table*}[t]
            \centering
            \begin{minipage}[t]{0.28\textwidth}
                \centering
                \vspace{0pt}
                \includegraphics[width=\linewidth]{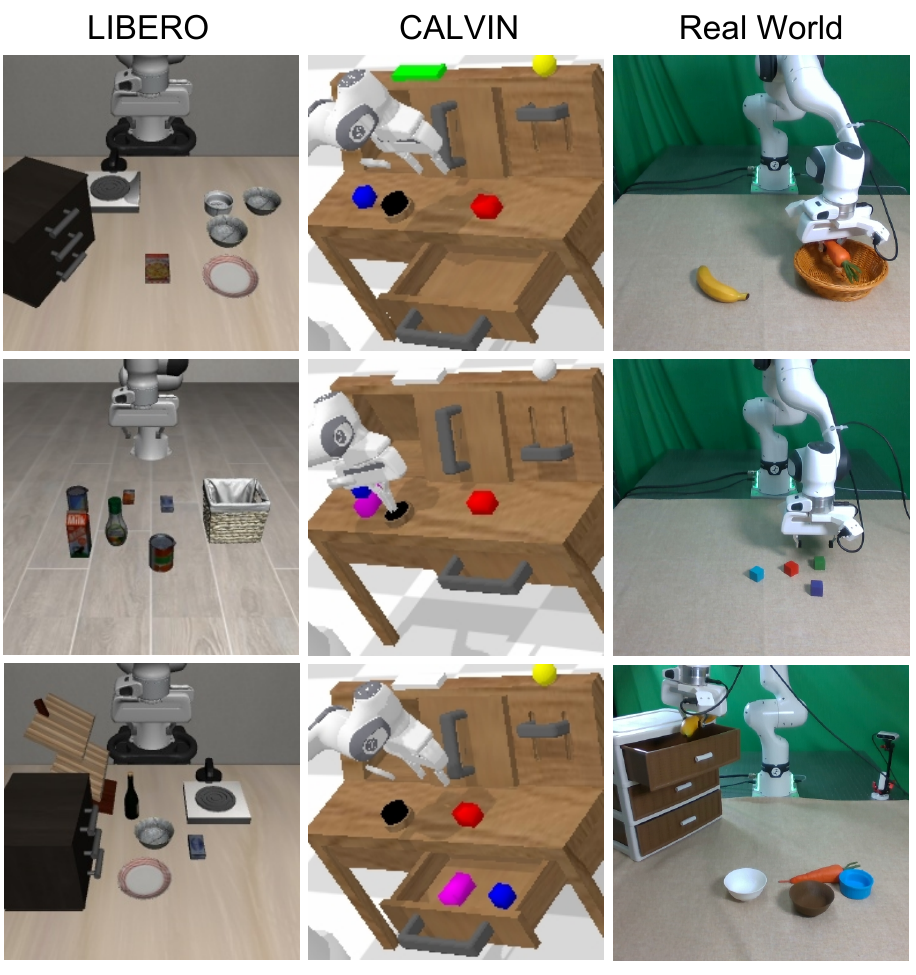}
                \captionof{figure}{Evaluation tasks}
                \label{fig:task}
            \end{minipage}
            \hfill
            \begin{minipage}[t]{0.70\textwidth}
                \centering
                \vspace{0pt}
                \captionof{table}{Results on CALVIN ABC$\rightarrow$D in success rates and average task completion length.}
                \vspace{1pt}
                \label{tab:calvin_main}
                \renewcommand{\arraystretch}{1.15}
                \setlength{\tabcolsep}{6pt}
                \resizebox{\linewidth}{!}{
                \begin{tabular}{lccccccc}
                    \toprule
                    \multirow{2}{*}[-0.3em]{Method} & \multirow{2}{*}[-0.3em]{Splits} & \multicolumn{5}{c}{Success Rate (\%)} & \multirow{2}{*}[-0.3em]{Avg. Len} \\
                    \cmidrule(lr){3-7}
                    & & 1/5 & 2/5 & 3/5 & 4/5 & 5/5 & \\
                    \midrule
                    Diffusion Policy         & ABC$\rightarrow$D & 40.2 & 12.3 & 2.6 & 0.8 & 0.0 & 0.56 \\
                    UniPi                    & ABC$\rightarrow$D & 56.0 & 16.0 & 8.0 & 8.0 & 4.0 & 0.92 \\
                    OpenVLA                  & ABC$\rightarrow$D & 91.3 &  77.8 & 62.0 & 52.1 & 43.5 & 3.27 \\
                    GR-1                     & ABC$\rightarrow$D & 85.4 & 71.2 & 59.6 & 49.7 & 40.1 & 3.06 \\
                    VPP                      & ABC$\rightarrow$D & \textbf{96.5} & 90.9 & \textbf{86.6} & \textbf{82.0} & 76.9 & \textbf{4.33} \\
                    \midrule
                    \rowcolor{gray!15}
                    \textbf{W2A} & ABC$\rightarrow$D & 95.3 & \textbf{91.7} & 85.5 & 80.2 & \textbf{77.1} & 4.30 \\
                    \bottomrule
                \end{tabular}
                }
            \end{minipage}
        \end{table*}
        
        \textbf{Real-world Tasks.} Table~\ref{tab:real_world_result} reports the average success rates on real-world tasks. W2A consistently outperforms both Diffusion Policy and OpenVLA across all three task types, achieving 78.3\%, 65.0\%, and 54.2\% success rates on pick-and-place, cube-stacking, and long-horizon tasks, respectively. The advantage becomes especially evident as task difficulty increases:
        \begin{table*}[t]
            \centering
            \caption{Results on real-world tasks (\% success rate).}
            \label{tab:real_world_result}
            \renewcommand{\arraystretch}{1.08}
            \begin{tabular}{lcccccc}
                \toprule
                Method & PnP & Cube & Long & Towel folding & Table clearing & USB insertion \\
                \midrule
                Diffusion Policy & 29.2 & 10.0 & 2.5 & 23.3 & 6.7 & 0.0 \\
                OpenVLA & 50.8 & 35.0 & 18.3 & 50.0 & 26.7 & 20.0 \\
                \midrule
                \rowcolor{gray!15}
                \textbf{W2A} & \textbf{78.3} & \textbf{65.0} & \textbf{54.2} & \textbf{76.7} & \textbf{63.3} & \textbf{50.0} \\
                \bottomrule
            \end{tabular}
        \end{table*}
        \begin{figure}[t]
            \centering
            \includegraphics[width=\linewidth]{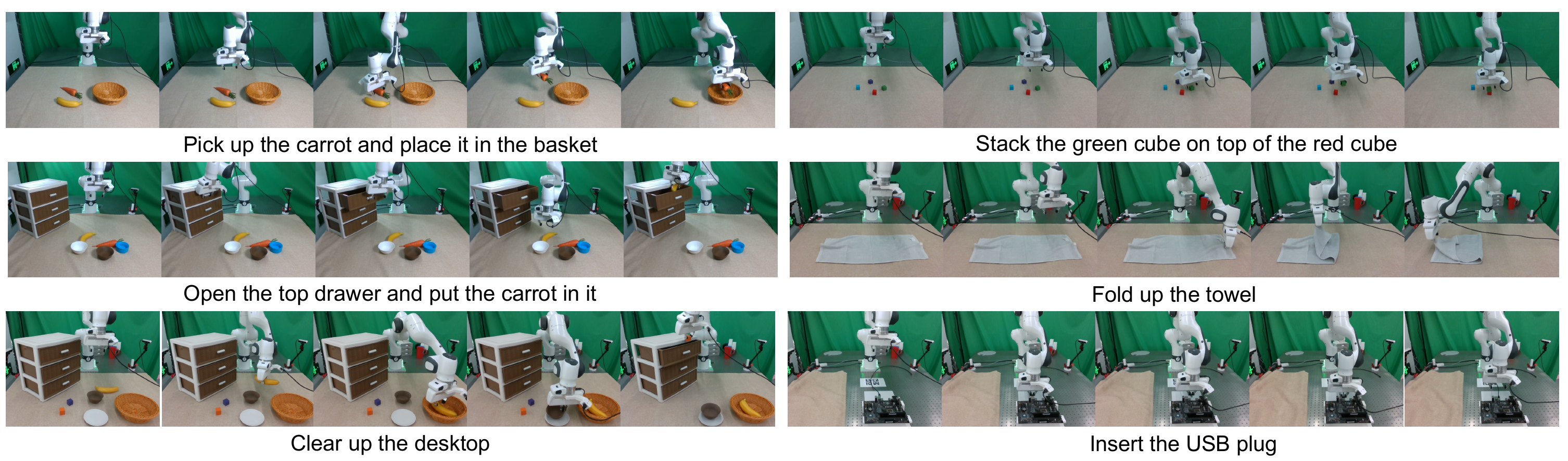}
            \caption{Representative real-world task cases.}
            \label{fig:task_case_simple}
        \end{figure}
        On the three additional practical tasks, W2A achieves 76.7\%, 63.3\%, and 50.0\% success rates on Towel folding, Table clearing, and USB insertion, respectively, compared with 50.0\%, 26.7\%, and 20.0\% for OpenVLA and 23.3\%, 6.7\%, and 0.0\% for Diffusion Policy. These results suggest that the distilled video priors help the action model capture more transferable scene dynamics and object interaction patterns, improving its robustness to variations in object instances, spatial configurations, and task goals. Figure~\ref{fig:task_case_simple} shows representative real-world task executions, covering pick-and-place, cube-stacking, long-horizon manipulation, and the three practical tasks. Detailed success rates for all subtasks and additional real-world task demonstrations are provided in Appendix~\ref{app:more_task_cases}.


        \begin{figure*}[t]
            \centering
            \begin{minipage}[t]{0.49\textwidth}
                \centering
                \includegraphics[width=\linewidth]{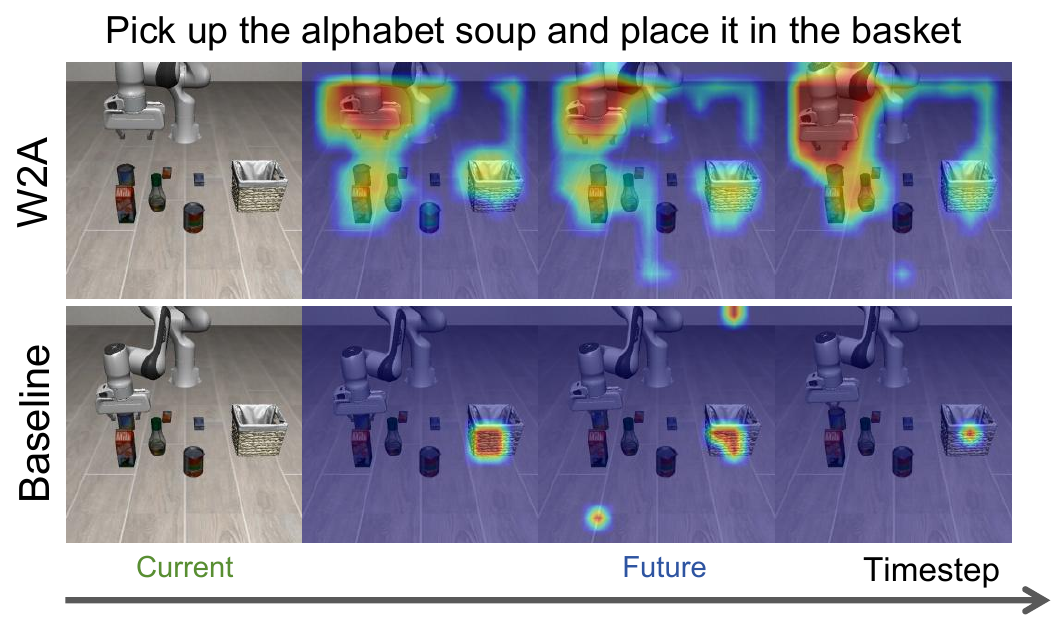}
                \caption{Similarity heatmap between action latent and teacher video latent.}
                \label{fig:sim_map}
            \end{minipage}
            \hfill
            \begin{minipage}[t]{0.49\textwidth}
                \centering
                \includegraphics[width=\linewidth]{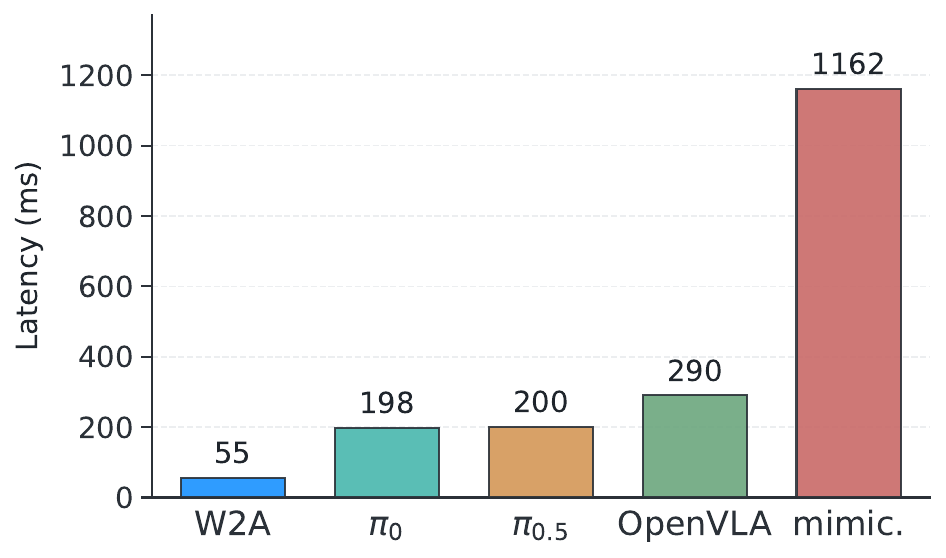}
                \caption{Inference latency comparison on the same hardware platform.}
                \label{fig:latency_bar}
            \end{minipage}
        \end{figure*}

        \begin{figure}[!t]
            \centering
            \includegraphics[width=\linewidth]{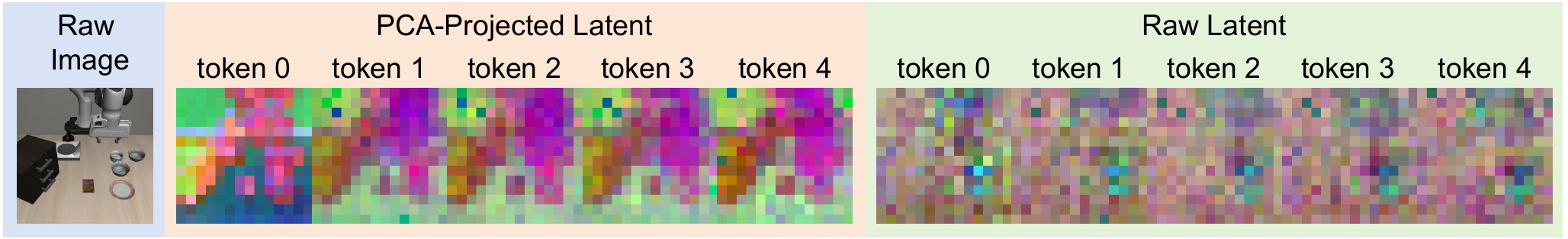}
            \caption{RGB visualizations of PCA-projected and raw teacher video latents across five temporal tokens, alongside the input image. PCA-projected latents show clearer spatial structure.}
            \label{fig:latent_rgb_short}
        \end{figure}
        
    \subsection{In-depth Analysis}
    \label{subsec:in-depth_analysis}
        \textbf{What is actually distilled?} We load the Q-Former as a probe at test time, visualize the similarity between the queried action latent and the teacher video latent with a heatmap, and compare it with an action DiT baseline trained without distillation. Figure~\ref{fig:sim_map} shows that the distilled action latent aligns more consistently with task-relevant regions in the teacher video latent (such as the gripper and manipulated objects), while background regions are largely ignored during distillation. Notably, the highlighted regions also correspond well to task-relevant areas in the future images, suggesting that the action DiT captures future-aware world priors rather than merely reflecting the current observation. In contrast, the heatmaps read out from the baseline do not exhibit consistent similarity patterns, indicating that W2A truly transfers latent knowledge from the video DiT into the action DiT, rather than merely allowing the Q-Former to fit the video latent. We provide additional experiments on mask-based task-relevant region alignment and linear probing in Appendix~\ref{app:world_prior_evidence} to further investigate the information distilled into the action latents.
    
        \textbf{What role does PCA play?} Figure~\ref{fig:latent_rgb_short} compares raw video latents and PCA-projected video latents. The PCA-projected latent preserves clearer scene layout and salient patterns in task-relevant regions, such as the gripper and target objects. In contrast, the raw latent visualization is noisier and less structured. This suggests that PCA preserves the dominant information in the teacher latent while suppressing redundant variations, providing a compact teacher representation for distillation. A more complete visualization and detailed discussion are provided in Appendix~\ref{app:visualization_and_discussion_of_raw_and_pca_projected_video_latents}.
        
        \textbf{Inference efficiency.} A key goal of W2A is to retain the benefit of video-based world priors without incurring the overhead of an explicit video model at test time. To verify this, we compare inference latency across representative baselines on the same hardware platform, a compute node equipped with a single NVIDIA A100 40GB GPU. As shown in Figure~\ref{fig:latency_bar}, W2A achieves the lowest average inference latency among all compared methods, requiring only 55\,ms per action prediction. In contrast, mimic-video, a WAM that retains explicit future imagination, incurs about $21\times$ higher latency. Moreover, with a lightweight text encoder, W2A requires only about 6 GB of GPU memory at deployment, which is an advantage in compute-constrained settings.

    \subsection{Ablations}
    \label{subsec:ablations}
        \begin{table}[htbp]
            \centering
            \caption{Ablation results on LIBERO, shown as SR \textcolor{mydeepred}{drops} (\%) relative to W2A.}
            \label{tab:ablation_libero}
            \renewcommand{\arraystretch}{1.15}
            \setlength{\tabcolsep}{5pt}
            \begin{tabular}{lccccc}
                \toprule
                \textbf{Method} & \textbf{Spatial} & \textbf{Object} & \textbf{Goal} & \textbf{Long} & \textbf{Avg.} \\
                \midrule
                w/o distill. & 85.9 (\textcolor{mydeepred}{-10.4}) & 87.7 (\textcolor{mydeepred}{-11.1}) & 83.6 (\textcolor{mydeepred}{-11.4}) & 80.5 (\textcolor{mydeepred}{-13.1}) & 84.4 (\textcolor{mydeepred}{-11.5}) \\
                w/o PCA & 87.0 (\textcolor{mydeepred}{-9.3}) & 92.3 (\textcolor{mydeepred}{-6.5}) & 90.2 (\textcolor{mydeepred}{-4.8}) & 84.8 (\textcolor{mydeepred}{-8.8}) & 88.6 (\textcolor{mydeepred}{-7.3}) \\
                \bottomrule
            \end{tabular}
        \end{table}
        We conduct ablation studies on two key design choices in W2A: the distillation signal and the PCA-based teacher compression. To this end, we compare W2A with two simplified variants: \emph{w/o distillation}, which removes the distillation objective and trains the action DiT only with the flow matching objective, and \emph{w/o PCA}, which directly uses the original teacher latent without feature compression. As shown in Table~\ref{tab:ablation_libero}, both variants perform worse than the full model. Removing distillation leads to a clear performance drop, confirming that the compact teacher video latents provide useful supervision beyond standard action modeling. Meanwhile, directly distilling the uncompressed teacher latent also degrades performance, suggesting that the raw teacher features are overly high-dimensional and noisy for effective transfer. These results show that both the proposed distillation design and PCA-based compression are important for transferring control-relevant world priors into the policy. More ablation results are provided in Appendix~\ref{app:supplementary_ablation_experiments}.

\section{Conclusion}
\label{sec:conclusion}
    In this paper, we present W2A, a two-stage distillation framework that transfers world priors from a video DiT into an action DiT for real-time robotic control. By compressing intermediate video latents and aligning them with action latents during training, W2A preserves useful video-based world priors while removing the video model entirely at test time. Experiments on LIBERO, CALVIN, and real-world tasks show that W2A achieves strong performance together with substantially lower inference latency. These results suggest that world priors learned by video models can be effectively distilled into action models for efficient real-time robot control.

\section{Limitations and Future Work}
\label{sec:limitation_and_future_work}
    Despite its strong efficiency-performance trade-off, W2A has several limitations. First, by distilling video-based world priors into an action model and removing the video model at test time, W2A gives up the additional model capacity and explicit future imagination ability of the video backbone. While crucial for real-time deployment, this design may limit robustness and generalization under large distribution shifts, unseen object dynamics, or tasks requiring longer-horizon adaptation beyond the distilled priors.

    Second, the framework compresses teacher latents into a fixed low-dimensional subspace before transfer to the student action model. This may not fully preserve task-relevant or higher-order structure in the original video latent space, limiting complex planning or broader generalization.

    A promising direction for future work is to explore more expressive teacher compression and distillation strategies that preserve richer world-model information while maintaining efficient deployment. We will also study hybrid designs that retain action-only inference while selectively invoking stronger world-model reasoning when needed, and extend the framework to more diverse embodiments, longer-horizon tasks, and open-world real-robot settings.

\clearpage
\acknowledgments{This work was supported by National Natural Science Foundation of China 62673160, Shenzhen Fundamental Research ZDCY20250901095402003, Guangdong Provincial Natural Science Foundation 2026A1515011727 and the Open Fund of Innovation Center for Control Actuators.}


\bibliography{ref}  

\newpage
\appendix

\section{Supplementary ablation experiments}
\label{app:supplementary_ablation_experiments}
    To further analyze the design choices in W2A, we conduct several supplementary ablation studies.
    
    \subsection{PCA K-sweep}
    \label{app:pca_k_sweep}
    
        \begin{figure*}[ht]
            \centering
            \begin{minipage}[t]{0.48\textwidth}
                \centering
                \includegraphics[width=\linewidth]{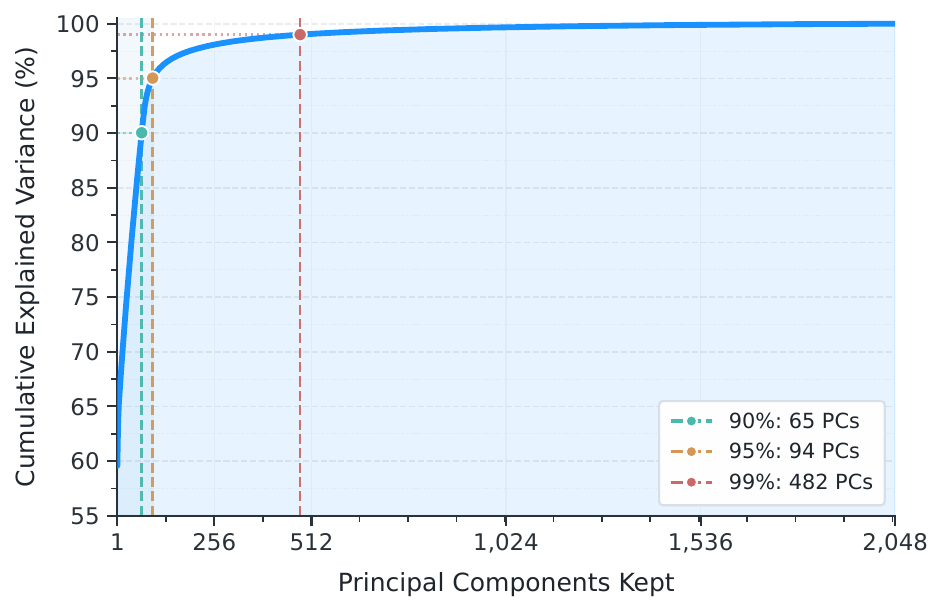}
                \caption{PCA cumulative explained variance curve of the teacher video latent.}
                \label{fig:pca_curve}
            \end{minipage}
            \hfill
            \begin{minipage}[t]{0.48\textwidth}
                \centering
                \includegraphics[width=\linewidth]{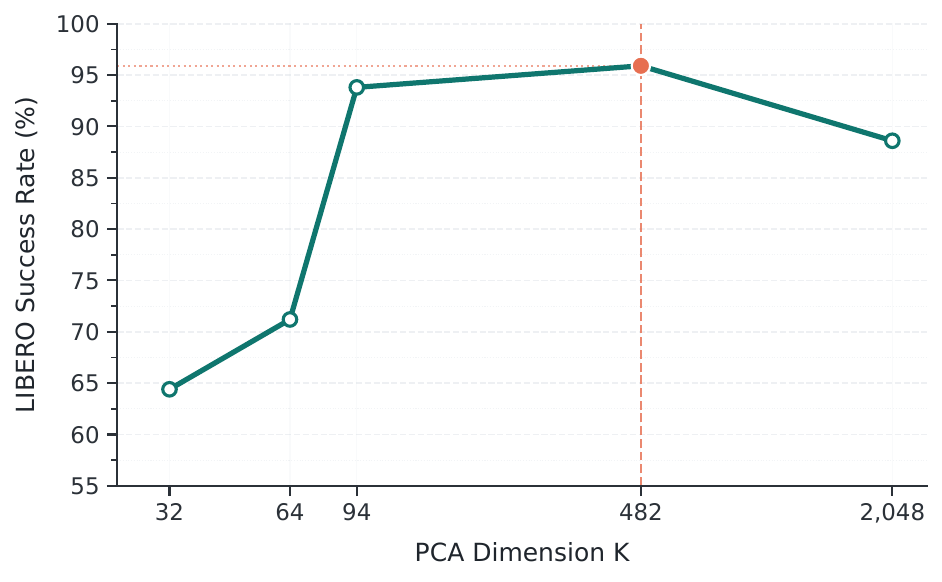}
                \caption{Results of PCA $K$-sweep experiment.}
                \label{fig:pca_k_success}
            \end{minipage}
        \end{figure*}
                
        We study the effect of the PCA compression dimension $K$ on distillation. Figure~\ref{fig:pca_curve} shows the cumulative explained variance ratio as a function of the PCA dimension. The curve increases rapidly at low dimensions and gradually saturates, indicating that a large portion of the variance in the teacher video latent is concentrated in a relatively compact subspace. In our experiments, we choose $K=482$, which corresponds to retaining approximately $99\%$ of the cumulative explained variance.
        
        We then evaluate how different PCA dimensions affect downstream policy performance on LIBERO. Figure~\ref{fig:pca_k_success} reports the success rate under different values of $K$. When $K$ is too small, the compressed teacher latent may discard useful control-relevant information, leading to weaker distillation. Increasing $K$ improves performance by preserving more information from the video latent. However, using an overly large $K$ does not necessarily bring further gains, as the retained dimensions may also include redundant or task-irrelevant visual variations, making the distillation target noisier for the action DiT. Overall, the results suggest that an intermediate compression dimension provides the best trade-off between preserving teacher information and filtering out irrelevant latent variations. This supports our choice of $K=482$, which preserves most of the teacher latent variance while keeping the distillation target compact.
    
    \subsection{Teacher Latent Depth Selection}
    \label{app:teacher_latent_depth_selection}
        \begin{wrapfigure}{r}{0.50\textwidth}
            \vspace{-35pt}
            \centering
            \includegraphics[width=\linewidth]{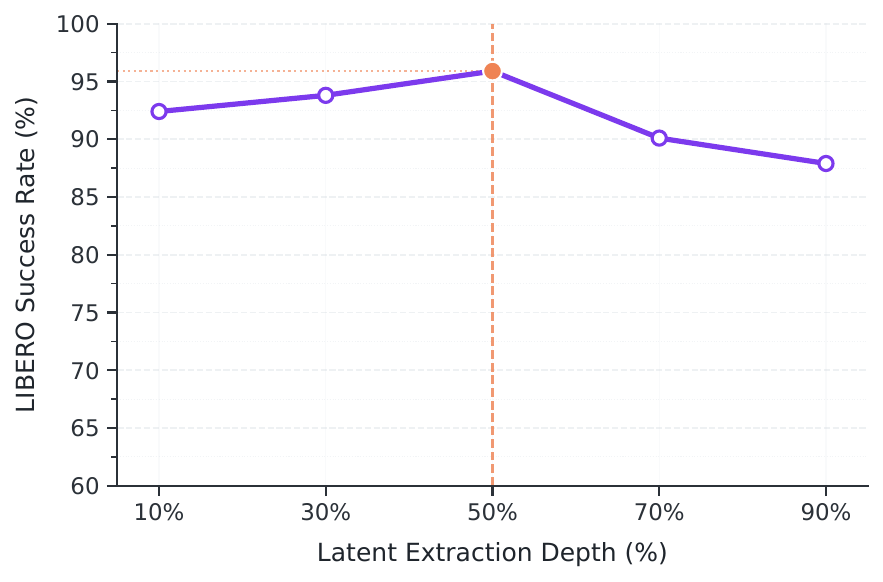}
            \caption{Effect of teacher latent extraction depth on LIBERO success rate.}
            \label{fig:teacher_depth}
            \vspace{-20pt}
        \end{wrapfigure}
        Teacher latent depth is an important factor in determining the quality of the distillation target. Different depths of the video DiT may encode different types of information: shallow representations tend to preserve more low-level visual details, while deeper representations may become more tied to the final video reconstruction objective. For action distillation, the teacher latent should retain useful spatio-temporal structure and task-relevant world priors, rather than being either too low-level or overly generation-specific.

        To analyze this effect, we extract teacher video latents from different relative depths of the video DiT while keeping the denoising time fixed. All other settings, including PCA compression, spatio-temporal alignment, and action DiT training, are kept unchanged. Figure~\ref{fig:teacher_depth} reports the LIBERO success rate under different latent extraction depths.
        
        The results show a clear non-monotonic trend. Using very shallow teacher latents leads to limited improvement, suggesting that early representations are not sufficiently informative for action-level distillation. Performance improves as the extraction depth increases and reaches the best result at the 50\% depth. However, further increasing the depth degrades performance, with the 90\% depth performing the worst. This indicates that overly deep video latents may be less suitable as distillation targets, possibly because they become more specialized toward video reconstruction rather than control-relevant structure. Based on this result, we use the intermediate teacher latent at 50\% depth as the default distillation target in W2A.

    \begin{table*}[h]
    \centering
        \caption{Ablation results on teacher signals on LIBERO, shown as success rate and performance \textcolor{mydeepred}{drops} (\%) relative to W2A.}
        \label{tab:ablation_teacher_signal}
        \renewcommand{\arraystretch}{1.15}
        \setlength{\tabcolsep}{5pt}
        \resizebox{\linewidth}{!}{
        \begin{tabular}{lcccccc}
            \toprule
            \textbf{Suite} & \textbf{w/o distill.} & \textbf{Shuffled Teacher} & \textbf{Random Teacher} & \textbf{Token0 Teacher} & \textbf{Current Feature} & \textbf{Future Feature} \\
            \midrule
            Spatial & 85.9 (\textcolor{mydeepred}{-10.4}) & 80.5 (\textcolor{mydeepred}{-15.8}) & 72.5 (\textcolor{mydeepred}{-23.8}) & 88.9 (\textcolor{mydeepred}{-7.4}) & 83.3 (\textcolor{mydeepred}{-13.0}) & 89.8 (\textcolor{mydeepred}{-6.5}) \\
            Object  & 87.7 (\textcolor{mydeepred}{-11.1}) & 81.6 (\textcolor{mydeepred}{-17.2}) & 79.8 (\textcolor{mydeepred}{-19.0}) & 91.0 (\textcolor{mydeepred}{-7.8}) & 86.6 (\textcolor{mydeepred}{-12.2}) & 92.5 (\textcolor{mydeepred}{-6.3}) \\
            Goal    & 83.6 (\textcolor{mydeepred}{-11.4}) & 77.7 (\textcolor{mydeepred}{-17.3}) & 65.2 (\textcolor{mydeepred}{-29.8}) & 84.4 (\textcolor{mydeepred}{-10.6}) & 81.7 (\textcolor{mydeepred}{-13.3}) & 86.4 (\textcolor{mydeepred}{-8.6}) \\
            Long    & 80.5 (\textcolor{mydeepred}{-13.1}) & 73.1 (\textcolor{mydeepred}{-20.5}) & 59.3 (\textcolor{mydeepred}{-34.3}) & 82.3 (\textcolor{mydeepred}{-11.3}) & 77.1 (\textcolor{mydeepred}{-16.5}) & 82.7 (\textcolor{mydeepred}{-10.9}) \\
            Avg.    & 84.4 (\textcolor{mydeepred}{-11.5}) & 78.2 (\textcolor{mydeepred}{-17.7}) & 69.2 (\textcolor{mydeepred}{-26.7}) & 86.7 (\textcolor{mydeepred}{-9.2}) & 82.2 (\textcolor{mydeepred}{-13.7}) & 87.8 (\textcolor{mydeepred}{-8.1}) \\
            \bottomrule
        \end{tabular}
        }
    \end{table*}

    \subsection{What Teacher Signal Matters?}
    \label{app:effect_of_the_istillation_signal}
        \subsubsection{Structured vs. Unstructured Teacher Signals}
            We study the role of the teacher video latent by comparing our full model with two variants: \emph{Shuffled Teacher}, which destroys the spatio-temporal ordering of teacher latents, and \emph{Random Teacher}, which replaces teacher latents with random targets of the same dimensionality. As shown in Table~\ref{tab:ablation_teacher_signal}, \emph{Shuffled Teacher} already leads to a substantial average drop of 17.7\%, with degradation across all four suites. This indicates that the gain of W2A depends not only on the presence of an auxiliary teacher signal, but also on preserving the original spatio-temporal organization of the teacher latent. In other words, the student does benefit from structured correspondence to the teacher video representation, rather than merely matching a bag of teacher features. At the same time, \emph{Random Teacher} causes an even larger average drop of 26.7\%, showing that the improvement is not explained by an additional regularization loss alone, but relies on meaningful non-random structure encoded in the video teacher. Overall, these results suggest that W2A benefits from both the informative content and the spatio-temporal organization of the teacher latent, with coherent teacher structure being particularly important for effective transfer.
        
        \subsubsection{Static, Future, and Video-DiT Teacher Latent}
            We further study whether the gain of W2A comes specifically from future-aware video teacher latents, rather than from static visual features or merely introducing an auxiliary teacher loss. To this end, we compare the default teacher source in W2A with three alternatives. \emph{Token0 Teacher} uses only the token corresponding to the current visual anchor from the video DiT, providing a primarily static teacher signal with limited future-aware dynamics. \emph{Current Feature} replaces the video teacher with ResNet features extracted from the current image, testing whether static visual representations alone are sufficient. \emph{Future Feature} uses ResNet features extracted from future frames, testing whether access to future visual information alone can account for the gain without relying on video-DiT latents.

            For fair comparison, all variants use the same student architecture, Q-Former readout, cosine distillation loss, and distillation weight, while only changing the teacher source. All teacher signals are projected to the same dimensionality and organized into the same temporal-spatial layout as the default teacher latent. Static teacher latents are repeated across temporal groups, while future-frame features are grouped using the same temporal compression setting as the default teacher. As shown in Table~\ref{tab:ablation_teacher_signal}, both static-feature teachers and future-frame teachers perform better than the unstructured variants, but all remain below the full model. In particular, \emph{Token0 Teacher} causes an average drop of 9.2\%, while \emph{Current Feature} leads to a larger average drop of 13.7\%, showing that static teacher signals alone are insufficient to explain the gain of W2A. \emph{Future Feature} is the strongest among the three additional variants, but still underperforms W2A by 8.1\% on average. This indicates that access to future visual information is helpful, yet does not fully match the benefit of the video-DiT teacher latent. These results suggest that the gain of W2A is not explained by static visual cues or auxiliary supervision alone, but is more closely related to the future-aware latent structure provided by the video DiT.

    \subsection{Effect of the Q-Former}
    \label{app:effect_of_the_qformer}
    \begin{wraptable}{r}{0.46\textwidth}
        \vspace{-35pt}
        \centering
        \captionsetup{width=\linewidth}
        \caption{Ablation results about Q-Former on LIBERO, shown as performance \textcolor{mydeepred}{drops} (\%) relative to W2A.}
        \label{tab:ablation_qformer}
        \renewcommand{\arraystretch}{1.08}
        \setlength{\tabcolsep}{3.5pt}
        \resizebox{\linewidth}{!}{
        \begin{tabular}{lcc}
            \toprule
            \textbf{Suite} & \textbf{Global MLP} & \textbf{Grouped MLP} \\
            \midrule
            Spatial & 83.7 (\textcolor{mydeepred}{-12.6}) & 89.5 (\textcolor{mydeepred}{-6.8}) \\
            Object  & 85.1 (\textcolor{mydeepred}{-13.7})  & 90.4 (\textcolor{mydeepred}{-8.4}) \\
            Goal    & 82.2 (\textcolor{mydeepred}{-12.8})  & 87.2 (\textcolor{mydeepred}{-7.8}) \\
            Long    & 79.8 (\textcolor{mydeepred}{-13.8}) & 83.8 (\textcolor{mydeepred}{-9.8}) \\
            Avg.    & 82.7 (\textcolor{mydeepred}{-13.2}) & 87.7 (\textcolor{mydeepred}{-8.2}) \\
            \bottomrule
        \end{tabular}
        }
        \vspace{-10pt}
    \end{wraptable}
    To assess whether the gain comes from the query-based readout itself or simply from introducing a learnable projector, we replace the Q-Former with two MLP-based projectors while keeping the teacher target and distillation loss unchanged. 
    The first is a \emph{Global MLP}, which flattens the entire action latent and directly predicts the aligned teacher latent grid. 
    The second is a more structured \emph{Grouped MLP}, which preserves the same temporal grouping as our full model: for each temporal group, a shared MLP maps the flattened action tokens to the corresponding teacher latent grid. 
    Both baselines produce the same output layout as the Q-Former model and are trained with the same cosine distillation loss.

    As shown in Table~\ref{tab:ablation_qformer}, both MLP-based projectors underperform the Q-Former. 
    The Global MLP performs the worst, suggesting that treating latent distillation as a monolithic regression problem is insufficient. 
    Preserving coarse temporal grouping with a Grouped MLP improves performance, but still falls short of the Q-Former. 
    This indicates that the gain of W2A does not come merely from adding a learnable projection head; rather, the query-based structured readout is more effective for extracting teacher-compatible features from action latents.

\section{Implementation Details}
\label{app:implementation}

    \subsection{Model Structure}
    \label{app:model_structure}
    
        Table~\ref{tab:model_config} summarizes the main architecture of W2A. The model consists of an observation encoder, an action DiT policy, and a training-only spatio-temporal alignment module for latent distillation. At inference time, the teacher video DiT and the alignment module are removed.
        
        \begin{table}[ht]
            \centering
            \caption{Main model configuration of W2A.}
            \label{tab:model_config}
            \renewcommand{\arraystretch}{1.12}
            \setlength{\tabcolsep}{6pt}
            \begin{tabular}{ll}
                \toprule
                \textbf{Component} & \textbf{Configuration} \\
                \midrule
                RGB observations & Front-view and wrist-view images \\
                Image resolution & $224 \times 224$ \\
                Vision encoder & ResNet-18~\citep{he2016deep} \\
                Observation history length & 2 \\
                Action horizon & 16 \\
                Executed horizon & 8 \\
                Language embedding & Precomputed T5 embedding~\citep{2020t5} \\
                \midrule
                Action DiT hidden size & 768 \\
                Action DiT depth & 10 transformer blocks \\
                Attention heads & 8 \\
                MLP ratio & 4.0 \\
                Condition injection & Cross-attention \\
                Flow-time injection & AdaLN \\
                \midrule
                Q-Former hidden dimension & 128 \\
                Q-Former layers & 1 \\
                Q-Former attention heads & 8 \\
                \bottomrule
            \end{tabular}
        \end{table}
        
        For latent distillation, we extract the student action representation from the middle layer of the action DiT and align it with a selected hidden state of the teacher video DiT. The main distillation configuration is shown in Table~\ref{tab:distill_config}.
        
        \begin{table}[ht]
            \centering
            \caption{Latent distillation configuration.}
            \label{tab:distill_config}
            \renewcommand{\arraystretch}{1.12}
            \setlength{\tabcolsep}{6pt}
            \begin{tabular}{ll}
                \toprule
                \textbf{Item} & \textbf{Configuration} \\
                \midrule
                Teacher latent & Output hidden state of the 14th video DiT block at $\tau_v=0.5$ \\
                Student latent & Output hidden state of the 5th action DiT block \\
                PCA dimension $K$ & 482 \\
                Teacher latent shape & $[B,24,14,14,2048]$ \\
                Teacher latent shape after PCA & $[B,24,14,14,482]$ \\
                Student latent shape & $[B,16,768]$ \\
                \bottomrule
            \end{tabular}
        \end{table}
    
    \subsection{Training Recipe}
    \label{app:training_recipe}
    
        \textbf{Training-data scope.} Throughout this paper, ``scratch'' and ``finetuned'' are distinguished solely by whether external robot-action data are used for policy pretraining. ``Scratch'' methods use robot-action data only from the target evaluation benchmark, whereas ``finetuned'' methods are pretrained on external robot-action data and then adapted to the target benchmark. This convention allows ``scratch'' methods to use generic video pretraining. In W2A, we adapt a pretrained video model using target-benchmark videos and train the action DiT from scratch using target-benchmark action data, without external robot-action pretraining.

        Table~\ref{tab:data_boundary} summarizes the training-data sources for representative methods. ``Ext. RA'' denotes external robot-action data, and ``Generic video'' denotes general video data. ``Target action'' and ``Target video'' refer to data from the target evaluation benchmark; target videos are counted only when explicitly used to train or adapt a video model. The ``Ext. RA'' column in Table~\ref{tab:libero_main} follows the same convention.

        \begin{table}[ht]
            \centering
            \caption{Training-data sources for representative methods. \cmark\ indicates use of the corresponding data source, and \xmark\ indicates no use.}
            \label{tab:data_boundary}
            \renewcommand{\arraystretch}{1.12}
            \setlength{\tabcolsep}{6pt}
            \resizebox{\linewidth}{!}{
            \begin{tabular}{llcccc}
                \toprule
                Category & Method & Ext. RA & Generic video & Target action & Target video \\
                \midrule
                Scratch & Diffusion Policy & \xmark & \xmark & \cmark & \xmark \\
                Scratch & mimic-video & \xmark & \cmark & \cmark & \cmark \\
                Scratch & DreamZero-5B & \xmark & \cmark & \cmark & \cmark \\
                Scratch & \textbf{W2A} & \xmark & \cmark & \cmark & \cmark \\
                \midrule
                Finetuned & Octo & \cmark & \xmark & \cmark & \xmark \\
                Finetuned & OpenVLA & \cmark & \xmark & \cmark & \xmark \\
                Finetuned & $\pi_{0.5}$ & \cmark & \xmark & \cmark & \xmark \\
                \bottomrule
            \end{tabular}
            }
        \end{table}

        Table~\ref{tab:training_config} summarizes the training recipe. We train the action DiT and the spatio-temporal alignment module jointly using the flow matching objective and the latent distillation objective.
        
        \begin{table}[ht]
            \centering
            \caption{Training configuration.}
            \label{tab:training_config}
            \renewcommand{\arraystretch}{1.12}
            \setlength{\tabcolsep}{6pt}
            \begin{tabular}{ll}
                \toprule
                \textbf{Item} & \textbf{Configuration} \\
                \midrule
                Batch size & 128 \\
                Training epochs & 100 \\
                Optimizer & AdamW \\
                Learning rate & $1\times10^{-4}$ \\
                Weight decay & $1\times10^{-4}$ \\
                Gradient clipping & 1.0 \\
                Learning-rate schedule & Cosine annealing \\
                EMA decay & 0.999 \\
                \bottomrule
            \end{tabular}
        \end{table}
        
        \begin{table}[ht]
            \centering
            \caption{Flow matching and distillation settings.}
            \label{tab:flow_config}
            \renewcommand{\arraystretch}{1.12}
            \setlength{\tabcolsep}{6pt}
            \begin{tabular}{ll}
                \toprule
                \textbf{Item} & \textbf{Configuration} \\
                \midrule
                Number of flow steps & 100 \\
                Flow-time sampling & Logit-normal \\
                Inference denoising steps & 20 \\
                Distillation weight $\lambda_{dist}$ & 1.0 \\
                Gaussian weighting center & $\tau_c=0.5$ \\
                Gaussian weighting std. & $\sigma_w=0.1$ \\
                \bottomrule
            \end{tabular}
        \end{table}

        \begin{table}[ht]
            \centering
            \caption{Inference configuration.}
            \label{tab:inference_config}
            \renewcommand{\arraystretch}{1.12}
            \setlength{\tabcolsep}{6pt}
            \begin{tabular}{ll}
                \toprule
                \textbf{Item} & \textbf{Configuration} \\
                \midrule
                Observation history & 2 steps \\
                Predicted action chunk & 16 actions \\
                Executed actions per prediction & 8 actions \\
                Inference denoising steps & 20 \\
                Warmstart time & $\tau_{\mathrm{ws}}=0.5$ \\
                \bottomrule
            \end{tabular}
        \end{table}
    
    \subsection{Inference Details}
    \label{app:inference_details}
    
        At inference time, W2A removes all training-only components, including the teacher video DiT and the spatio-temporal alignment module. The deployed policy only consists of the observation encoder, language embedding, action DiT, and action normalizer. By default, we use precomputed T5 embeddings as the text input. For measuring inference latency and GPU memory usage, we instead use the variant trained with a CLIP text encoder. The two variants achieve comparable policy performance.
        
        After each prediction, only the first 8 actions are executed, and the unexecuted suffix is cached to warmstart the next prediction. This enables temporally consistent replanning while keeping inference video-free. Table~\ref{tab:flow_config} and Table~\ref{tab:inference_config} provide the configurations for flow matching and inference, respectively.
    
    \subsection{Spatio-temporal Alignment Mechanism}
    \label{app:temporal_spatial_alignment}
        We provide additional details of the spatio-temporal alignment mechanism used in Stage II. The goal is to organize the Q-Former output into the same spatio-temporal layout as the compressed teacher video latent $\tilde{Z}^v$, enabling position-wise distillation between corresponding spatio-temporal locations. In our current implementation, the Q-Former mainly serves as a structured information readout module.

        In our implementation, $H=16$ and $D_a=768$. Following the setting used in our LoRA adaptation of Cosmos-Predict2, the video VAE has a temporal compression ratio of $r_t=4$, meaning that one video temporal token summarizes four future frames. Accordingly, we partition $Z^a$ into $G=H/r_t$ temporal groups:
        \begin{equation}
            Z^a \rightarrow \{Z^a_g\}_{g=1}^{G},
            \qquad
            Z^a_g \in \mathbb{R}^{r_t \times D_a}.
        \end{equation}
        With $H=16$ and $r_t=4$, we obtain $G=4$ temporal groups, each containing four action tokens.
        
        On the teacher side, we exclude the temporal token corresponding to the current observation and use the following $G$ selected video temporal tokens in $\tilde{Z}^{v}$ as distillation targets. For each temporal group $g$, we introduce a set of learnable spatial query tokens arranged according to the spatial resolution of $\tilde{Z}^{v}$. These queries aggregate information from the action tokens within the same temporal group through cross-attention:
        \begin{equation}
            \hat{Z}^a_g = \mathrm{CrossAttn}(Q_g, Z^a_g),
        \end{equation}
        where $Q_g$ denotes the spatial queries for the $g$-th temporal group. This produces one aligned action representation for each spatial position of the compressed teacher video latent.
        
        Applying this operation across all temporal groups yields
        \begin{equation}
            \hat{Z}^a \in \mathbb{R}^{G \times H_v W_v \times K},
        \end{equation}
        which matches the layout of the selected compressed teacher latent:
        \begin{equation}
            \tilde{Z}^v_{\mathrm{sel}} \in \mathbb{R}^{G \times H_v W_v \times K}.
        \end{equation}
        Both tensors are then flattened into $[G H_v W_v, K]$ for computing the cosine distillation loss. This flattening is only used for loss computation and preserves the correspondence between the same temporal group and the same spatial position.





\section{Visualization and Discussion of Raw and PCA-Projected Video Latents}
\label{app:visualization_and_discussion_of_raw_and_pca_projected_video_latents}
\begin{figure*}[t]
    \centering
    \includegraphics[width=\textwidth]{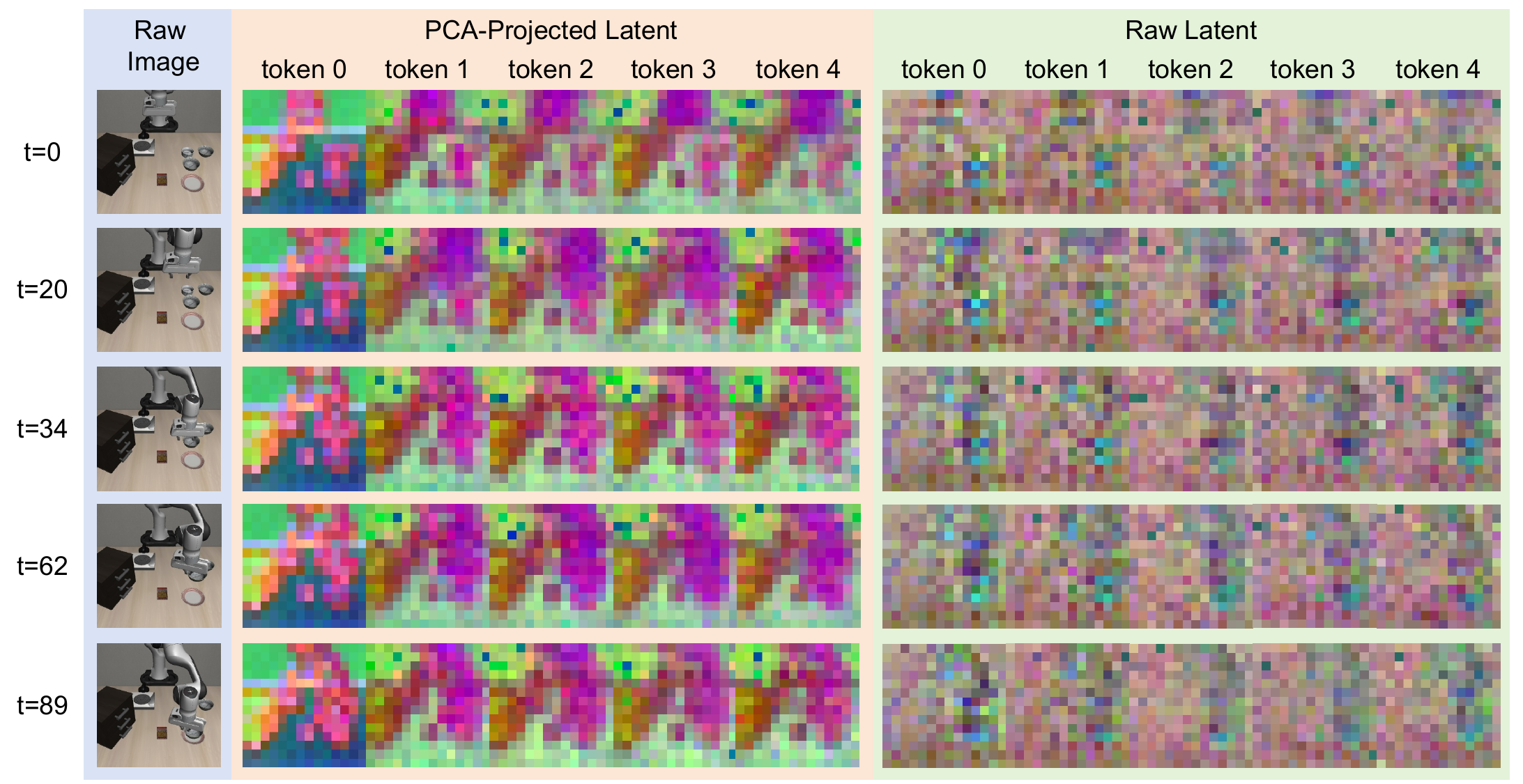} 
    \caption{Comparison of raw and PCA-projected video latents visualized as RGB feature maps.}
    \label{fig:latent_rgb}
\end{figure*}
Figure~\ref{fig:latent_rgb} compares raw video latents and PCA-projected video latents. For visualization, we take the first three channels as RGB-like values, apply a symmetric normalization, and map them to $[0,255]$. For the PCA-projected latent, these three channels correspond to the first three principal directions, while for the raw latent they correspond to the first three original feature channels.

The PCA-projected latent remains visually more structured and interpretable than the raw latent. In particular, it preserves clear scene layout and salient patterns in dynamic and task-relevant regions, such as the gripper and target objects. By contrast, the raw latent visualization is noisier and less structured. This suggests that PCA preserves the dominant information in the teacher latent while suppressing part of the redundant variations, which is consistent with our use of PCA to construct a compact teacher representation for distillation.

This also provides additional qualitative evidence that the compressed teacher video latent retains scene dynamics and task-relevant interaction cues that are transferred from the video DiT to the action DiT.

\begin{figure*}[htbp]
    \centering
    \includegraphics[width=0.9\textwidth]{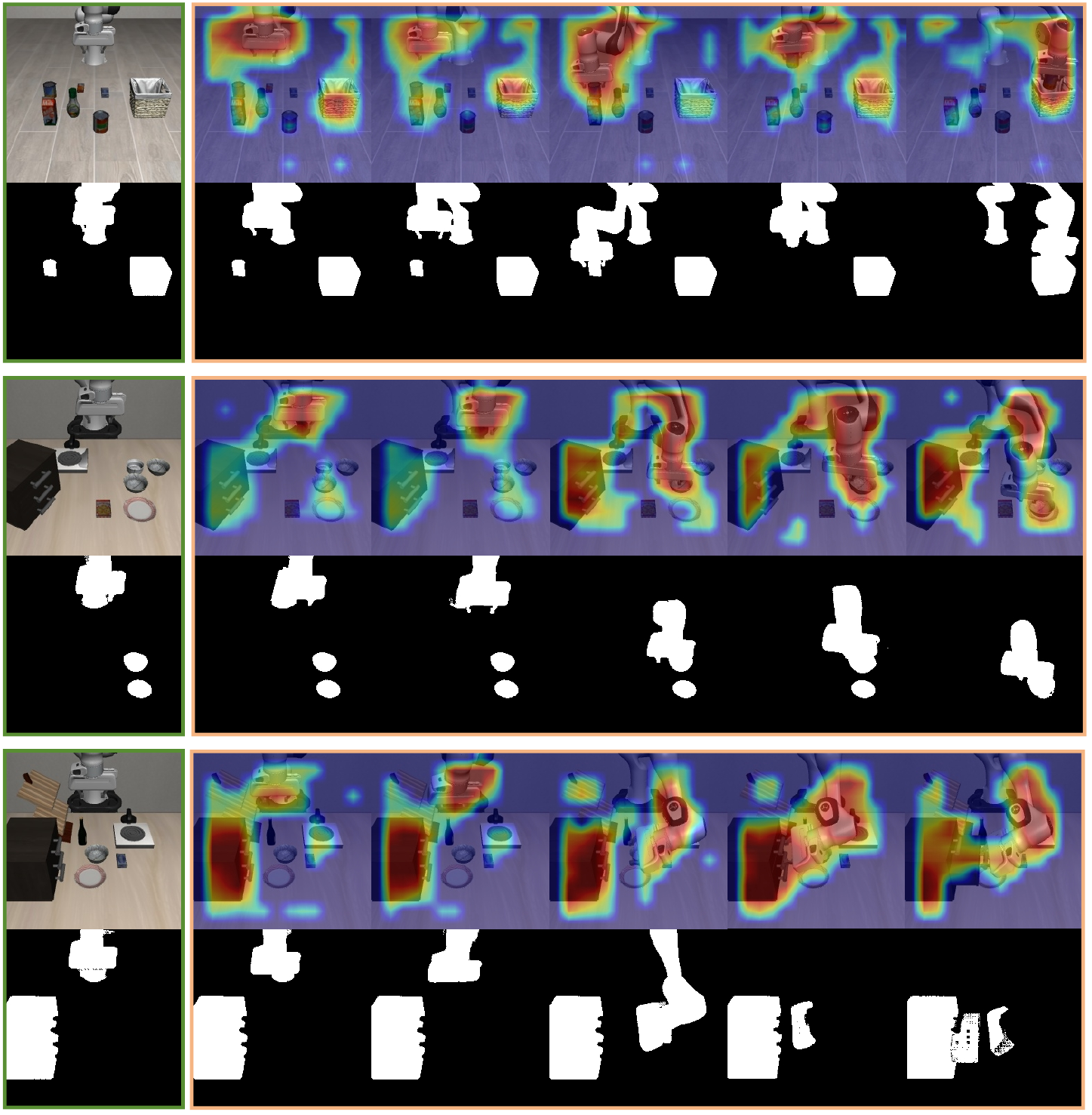} 
    \caption{Examples of task-relevant masks and action-video alignment maps used for world-prior analysis. SAM2 provides frame-level masks, and warmer colors denote higher latent similarity.}
    \label{fig:sim_and_mask}
\end{figure*}
\section{Quantitative Evidence of Distilled Task-Relevant World Priors}
\label{app:world_prior_evidence}
\subsection{Task-Relevant Region Alignment}
As a quantitative extension to the analysis of distilled world priors in Sec.~\ref{subsec:in-depth_analysis}, we further measure whether the action representations shaped by W2A are more strongly aligned with task-relevant regions of the teacher video latent. Since the spatial components of world priors that are most relevant to manipulation should primarily involve the gripper and manipulated objects, we compute region-level statistics over the action-video latent similarity maps. Specifically, we use Segment Anything 2 (SAM2)~\citep{ravi2024sam2} to automatically segment object masks from the corresponding video frames, and select the masks associated with the task-relevant objects. These masks are then downsampled to the $H_v\times W_v$ teacher latent grid using area interpolation, resulting in a soft mask $M\in[0,1]^{H_v\times W_v}$. The action-video alignment is measured using the same aligned-slot cosine similarity map as in Sec.~\ref{subsec:in-depth_analysis}. Figure~\ref{fig:sim_and_mask} shows examples of the SAM2-derived masks and the corresponding action-video alignment maps used in this analysis.

We report two region-level metrics. The first is \emph{mask enrichment}, which measures how much similarity mass falls inside the task-relevant mask after normalizing by the mask area. Given a similarity map $S$, we convert it into a spatial probability distribution $P$ using softmax normalization, i.e., $P_i=\exp(S_i/\tau_s)/\sum_j\exp(S_j/\tau_s)$, and compute
\[
E_{\mathrm{mask}}=\frac{\sum_i M_i P_i}{\frac{1}{N}\sum_i M_i+\epsilon},
\]
where $i$ indexes spatial locations, $N=H_vW_v$, and $\tau_s$ is the softmax temperature. A uniform similarity distribution has an expected enrichment of $1.0\times$, while values larger than $1.0\times$ indicate that the action-video latent similarity is concentrated in task-relevant regions beyond what would be expected from their area alone. The second metric is \emph{mask-background contrast}, which measures whether the mask region has higher normalized similarity than the background. We first standardize the similarity map as $\hat{S}_i=(S_i-\mu_S)/(\sigma_S+\epsilon)$, and then compute
\[
C_{\mathrm{mask-bg}}=
\frac{\sum_i M_i \hat{S}_i}{\sum_i M_i+\epsilon}
-
\frac{\sum_i (1-M_i)\hat{S}_i}{\sum_i (1-M_i)+\epsilon}.
\]
A larger contrast indicates that the task-relevant region is more strongly aligned with the teacher video latent than the surrounding background.

\begin{table}[t]
\centering
\caption{Quantitative evidence of distilled task-relevant world priors. We measure whether action-video latent similarity is concentrated in task-relevant object regions. Mask enrichment is area-normalized, where a uniform similarity map has an expected value of $1.0\times$. Mask-bg contrast is computed on standardized similarity maps.}
\label{tab:world_prior_quant}
\begin{tabular}{lcc}
\toprule
Method & Mask enrichment $\uparrow$ & Mask-bg contrast $\uparrow$ \\
\midrule
Action DiT w/o distill. & 0.98$\times$ & 0.01 \\
W2A & \textbf{3.87$\times$} & \textbf{1.34} \\
\bottomrule
\end{tabular}
\end{table}

As shown in Table~\ref{tab:world_prior_quant}, the non-distilled Action DiT baseline exhibits only marginal alignment with the task-relevant mask, with a mask enrichment close to the uniform-distribution baseline ($0.98\times$) and an almost zero mask-background contrast ($0.01$). In contrast, W2A achieves a substantially higher mask enrichment of $3.87\times$ and a much stronger mask-background contrast of $1.34$. This indicates that the action-video latent similarity in W2A is not uniformly distributed over the scene, but is strongly concentrated in task-relevant object regions. These results provide quantitative evidence that W2A encourages the action model to internalize task-relevant world priors from the video DiT, especially those associated with manipulated objects and their interactions, rather than merely imposing a generic feature-level regularization.

\subsection{Linear Probe Test}
\label{app:linear_probe}
To further test whether future dynamics are encoded in the action representation, we freeze the action DiT and train a linear probe to predict the future gripper and object image-plane coordinates from its action latents. We evaluate predictions at the current step $t$ and at future offsets $t{+}5$ and $t{+}10$. We report the positional endpoint error (Pos EPE) in pixels, where lower values indicate more accurate prediction. The ``w/o Distill'' baseline uses the same action model without the latent distillation objective.

\begin{table}[t]
    \centering
    \caption{Results of the linear probe test for predicting future gripper and object image-plane coordinates. Pos EPE is measured in pixels ($\downarrow$).}
    \label{tab:world_prior_linear_probe}
    \begin{tabular}{lccc}
        \toprule
        Method & \multicolumn{3}{c}{Pos endpoint error (EPE) (px) $\downarrow$} \\
        \cmidrule(lr){2-4}
        & $t$ & $t{+}5$ & $t{+}10$ \\
        \midrule
        Action DiT w/o distill. & 6.4 & 13.7 & 23.8 \\
        \textbf{W2A} & \textbf{5.9} & \textbf{8.2} & \textbf{13.1} \\
        \bottomrule
    \end{tabular}
\end{table}

As shown in Table~\ref{tab:world_prior_linear_probe}, W2A achieves lower Pos EPE at all prediction horizons than the non-distilled action model. The improvement becomes larger for farther future steps, suggesting that the distilled action representation retains information about how the gripper and manipulated objects evolve over time. This provides complementary evidence that W2A learns future-aware world priors rather than only improving instantaneous action prediction.

\section{Discussion of the Q-Former and Latent Space Distillation}
\label{app:disccusion_of_qformer_and_latent_space_distillation}
W2A is better viewed as a form of \emph{intermediate representation distillation} rather than classical response-based distillation. While early knowledge distillation focused on soft outputs~\citep{Hinton2015DistillingTK}, later work showed that useful teacher knowledge can also be transferred through hidden features, attention, and relational structure~\citep{Romero2014FitNetsHF,Zagoruyko2017AT,park2019relational,Jiao2019TinyBERTDB}. From this perspective, our method distills compact video-side latent representations into an action model, and therefore naturally belongs to the broader family of representation distillation.

The role of Q-Former in W2A should also be interpreted in light of our implementation. In BLIP-2, Q-Former is introduced as a lightweight querying transformer that uses learnable queries to extract useful information from frozen visual tokens~\citep{li2023blip}. More broadly, multimodal systems often employ lightweight bridge modules to connect pretrained backbones and downstream objectives~\citep{Alayrac2022FlamingoAV,wang2024vilaefficientvideolanguagealignment}. In our case, the Q-Former mainly serves as a structured information readout module that organizes action latents into teacher-compatible slots, which is consistent with our implementation details in Appendix~\ref{app:temporal_spatial_alignment}. This design is also compatible with recent evidence that generative and diffusion models provide transferable intermediate representations that can be distilled into downstream models~\citep{Yang2023DiffusionMA,Deng2026RobotDIFTDD}.

\section{Hardware}
\label{app:hardware}
    \begin{figure*}[t]
        \centering
        \includegraphics[width=\textwidth]{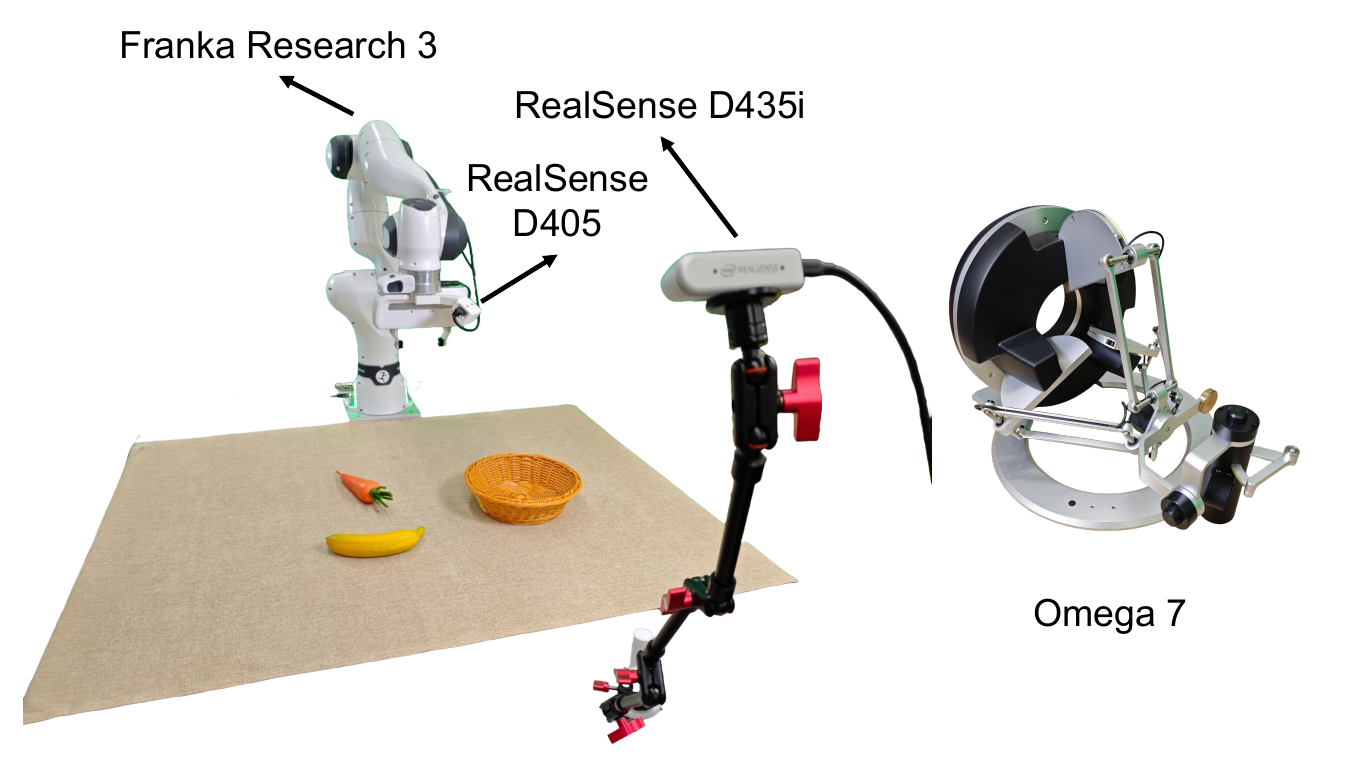} 
        \caption{Real-world hardware setup. We use a Franka Research 3 robot for policy execution and an Omega 7 haptic device for teleoperated demonstration collection.}
        \label{fig:hardware}
    \end{figure*}
    Figure~\ref{fig:hardware} shows the hardware used in our real-world experiments. We use a Franka Research 3 robot as the manipulation platform. The robot is equipped with a front-view camera and a wrist-view camera, which provide the visual observations used by the policy. The policy runs at 15 Hz on the real robot and predicts action chunks in a receding-horizon manner.

    For data collection, we use an Omega 7 haptic device to teleoperate the robot and collect demonstrations. The operator controls the robot end effector through the Omega 7 device, while the robot observations and actions are recorded to construct the real-world training dataset. This setup is used for both pick-and-place and cube-stacking tasks.

\section{Detailed Simulation Results Across Random Seeds}
\label{app:detailed_simulation_results}
    \begin{table}[t]
        \centering
        \caption{Results across different random seeds on LIBERO and CALVIN.}
        \label{tab:seed_results}
        \renewcommand{\arraystretch}{1.05}
        \setlength{\tabcolsep}{6pt}
        \begin{tabular}{ccccc|c}
            \toprule
            \multirow{3}{*}{Seed} & \multicolumn{4}{c|}{LIBERO} & CALVIN \\[-1.5ex]
            & \multicolumn{4}{c|}{\rule{8cm}{0.4pt}} & \multicolumn{1}{c}{\rule{1.6cm}{0.4pt}} \\
            & Spatial & Object & Goal & Long & Avg. Len \\
            \midrule
            0   & 95.4 & 99.0 & 91.6 & 91.0 & 4.34 \\
            5   & 98.6 & 98.2 & 97.4 & 94.2 & 4.26 \\
            42  & 95.0 & 99.2 & 96.0 & 95.6 & 4.30 \\
            \midrule
            Avg. & 96.3$\pm$1.97 & 98.8$\pm$0.53 & 95.0$\pm$3.03 & 93.6$\pm$2.36 & 4.30$\pm$0.04 \\
            \bottomrule
        \end{tabular}
    \end{table}
    
    \begin{figure*}[t]
        \centering
        \includegraphics[width=0.6\textwidth]{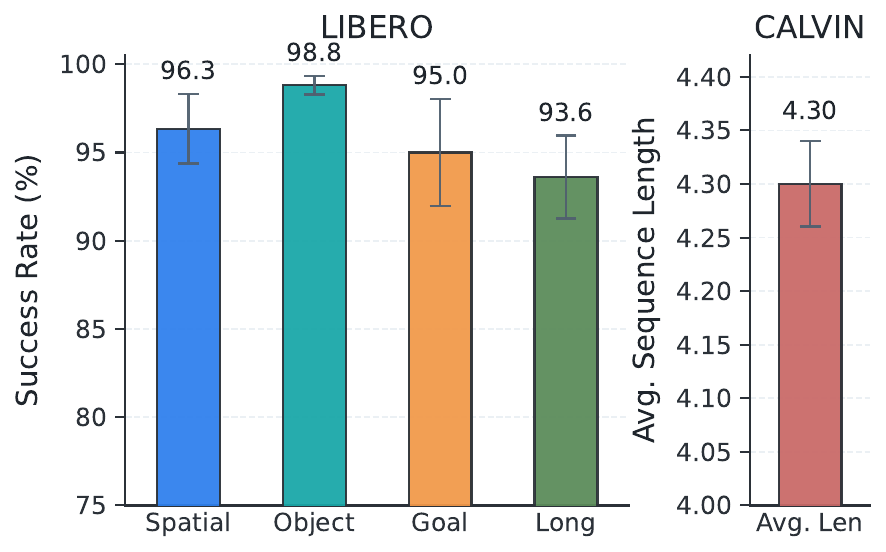} 
        \caption{Seed-wise simulation results. Error bars denote standard deviation over three seeds.}
        \label{fig:detailed_results_bar}
    \end{figure*}

    As shown in Table~\ref{tab:seed_results} and Figure~\ref{fig:detailed_results_bar}, W2A exhibits stable performance across different random seeds on both LIBERO and CALVIN. The Object suite shows the smallest variance across seeds, indicating highly stable performance on object-centric manipulation tasks. The Goal and Long suites exhibit relatively larger variation, which is expected since they involve more diverse goal configurations and longer-horizon execution. Nevertheless, W2A maintains strong performance under all three seeds, suggesting that the reported gains are not caused by a favorable random initialization.

    On CALVIN ABC$\rightarrow$D, W2A also shows stable long-horizon instruction-following performance, with an average sequence length of 4.30$\pm$0.04 across three seeds. The small variance indicates that W2A generalizes consistently to the unseen environment split. Overall, these results demonstrate that W2A achieves not only high average performance but also robust behavior across random seeds.

\section{Detailed Real-world Results and More Demonstrations}
\label{app:more_task_cases}
    \begin{table*}[tp]
        \centering
        \caption{Per-subtask results on real-world tasks. Each entry is reported as the number of successful rollouts out of 30. The overall row reports the average success rate over all subtasks.}
        \label{tab:real_world_subtasks}
        \renewcommand{\arraystretch}{1.15}
        \setlength{\tabcolsep}{6pt}
        \resizebox{\linewidth}{!}{
        \begin{tabular}{ccccc}
            \toprule
            Task Type & Subtask & Diffusion Policy & OpenVLA & W2A \\
            \midrule
    
            \multirow{4}{*}{PnP}
            & Pick up the carrot and place it in the basket & 12/30 & 15/30 & 25/30 \\
            & Pick up the banana and place it in the basket & 6/30 & 15/30 & 28/30 \\
            & Pick up the white bowl and place it on the plate & 9/30 & 22/30 & 19/30 \\
            & Pick up the blue cup and place it on the plate & 8/30 & 9/30 & 22/30 \\
            \midrule
    
            \multirow{4}{*}{Cube}
            & Stack the green cube on top of the red cube & 0/30 & 10/30 & 21/30 \\
            & Stack the purple cube on top of the red cube & 4/30 & 10/30 & 15/30 \\
            & Stack the green cube on top of the purple cube & 8/30 & 6/30 & 27/30 \\
            & Stack the red cube on top of the blue cube & 0/30 & 16/30 & 15/30 \\
            \midrule
    
            \multirow{4}{*}{Long}
            & Open the top drawer and put the carrot in it & 1/30 & 6/30 & 21/30 \\
            & Open the top drawer and put the banana in it & 1/30 & 3/30 & 14/30 \\
            & Open the top drawer and put the white bowl in it & 1/30 & 0/30 & 10/30 \\
            & Pick up the carrot and banana and place them in the basket & 0/30 & 13/30 & 20/30 \\
            \midrule

            \multirow{3}{*}{Practical Skills}
            & Towel folding & 7/30 & 15/30 & 23/30 \\
            & Table clearing & 2/30 & 8/30 & 19/30 \\
            & USB insertion & 0/30 & 6/30 & 15/30 \\
            \midrule
    
            \rowcolor{gray!15}
            \multicolumn{2}{c}{\textbf{Overall}} & 13.1\% & 34.2\% & 65.3\% \\
            \bottomrule
        \end{tabular}
        }
    \end{table*}

    \begin{figure*}[t]
        \centering
        \includegraphics[width=\textwidth]{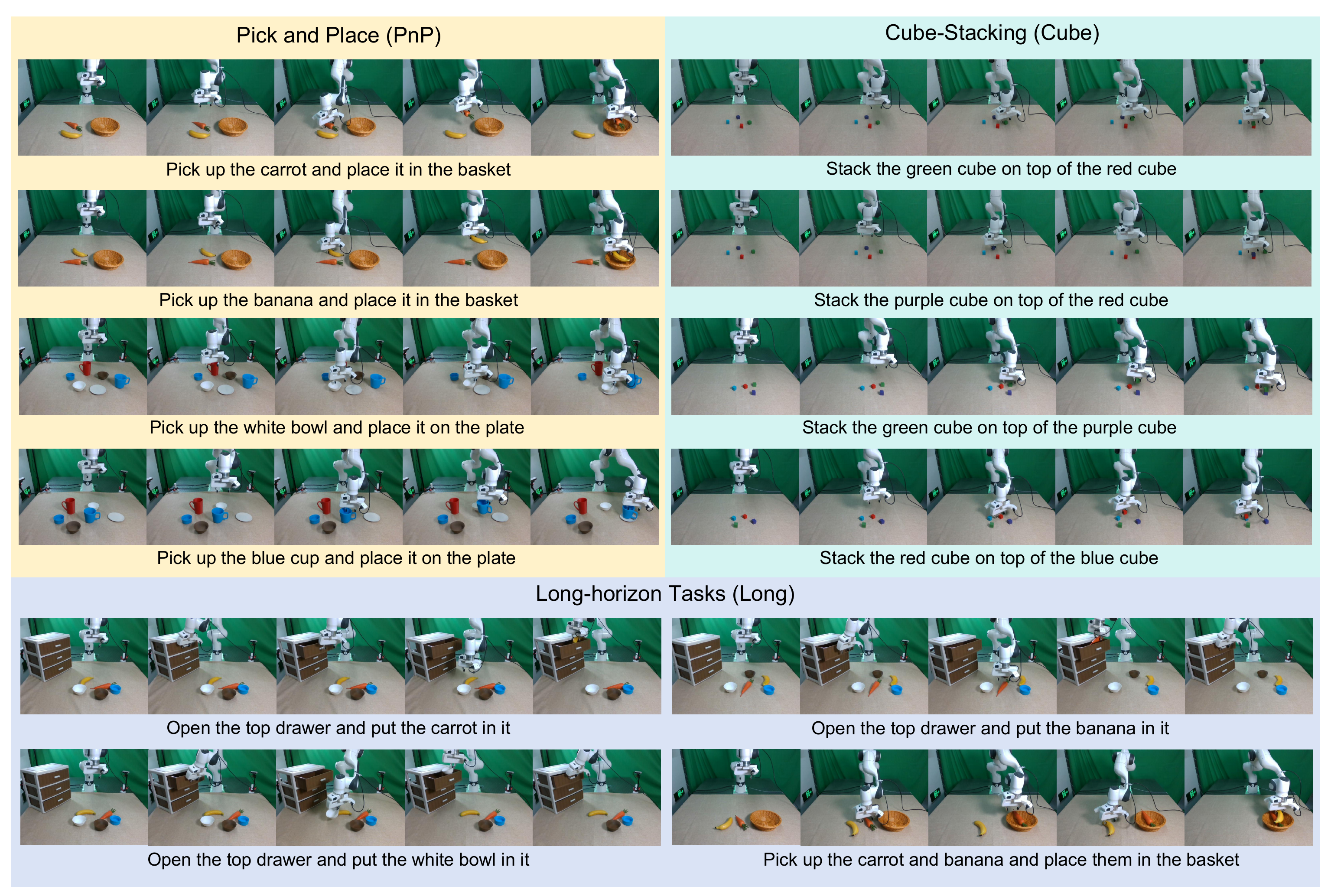} 
        \caption{Additional real-world task cases.}
        \label{fig:task_case}
    \end{figure*}
    Figure~\ref{fig:task_case} shows additional real-world task cases used in our evaluation. The examples cover both pick-and-place and cube-stacking tasks. For pick-and-place, the robot needs to identify the target object from language instructions and place it into the specified receptacle or onto the target object. For cube-stacking, the robot must manipulate small objects and complete sequential placement under contact-rich interactions. Detailed per-subtask real-world results are reported in Table~\ref{tab:real_world_subtasks}, where each entry shows the number of successful rollouts out of 30.

\section{Generalization Behavior and Failure Mode Analysis}
\label{app:generalization_failure}
    \begin{figure*}[t]
        \centering
        \includegraphics[width=\textwidth]{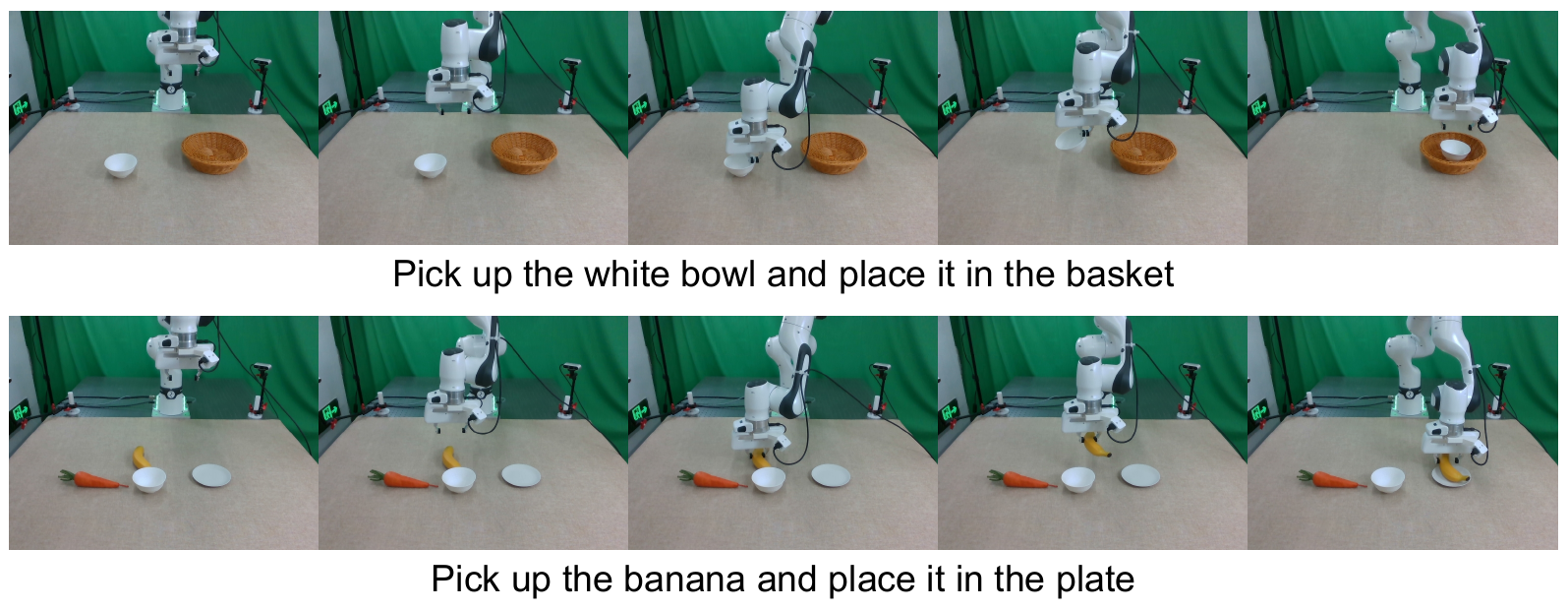} 
        \caption{Examples of compositional generalization in real-world evaluation.}
        \label{fig:generalization_cases}
    \end{figure*}
    \begin{figure*}[t]
        \centering
        \includegraphics[width=\textwidth]{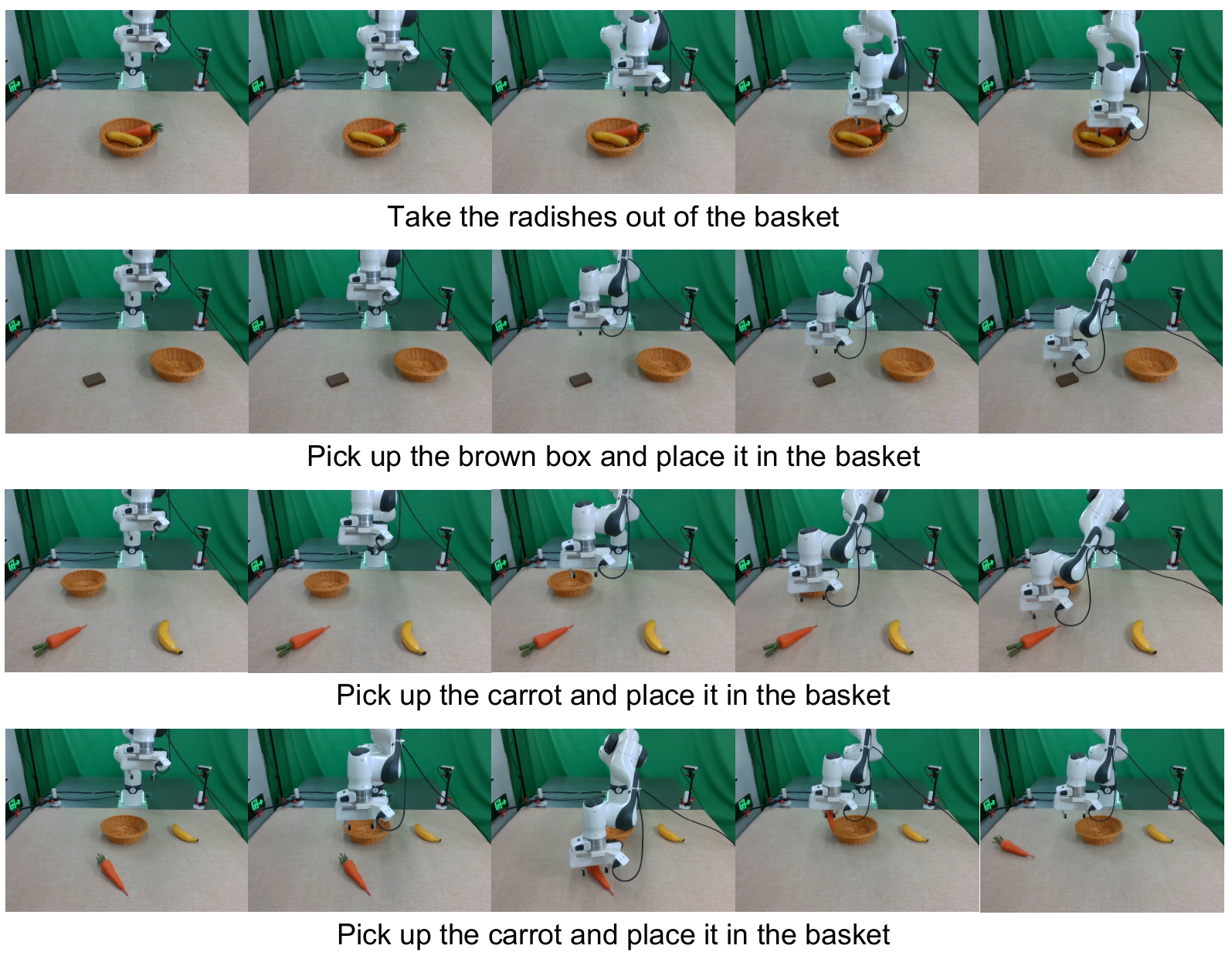} 
        \caption{Representative failure cases in real-world evaluation. W2A may fail when facing unseen tasks or objects, or when the initial scene configuration deviates substantially from the training distribution.}
        \label{fig:failure_cases}
    \end{figure*}
    We further examine the compositional generalization behavior of W2A in real-world evaluation. As shown in Figure~\ref{fig:generalization_cases}, W2A can complete new combinations of previously observed objects, receptacles, and manipulation skills, such as placing the white bowl into the basket and placing the banana onto the plate. These rollouts are not direct repetitions of training demonstrations: while the involved objects and primitive skills have been seen during training, the specific object-task pairings and resulting motion trajectories are unseen. This suggests that W2A can recombine learned interaction patterns and task-conditioned motion priors, rather than simply memorizing demonstrated trajectories.

    However, this compositional generalization is not unlimited. When the task semantics or scene configuration move beyond the support of the training distribution, W2A can still fail, as shown in Figure~\ref{fig:failure_cases}. These failures mainly arise in two situations. First, W2A may fail on tasks or objects that are entirely unseen during training, such as novel object categories or unseen task specifications. In these cases, the policy cannot directly rely on previously learned object interaction patterns or task-conditioned motion priors. Second, W2A can also fail when the initial scene configuration differs too much from the training distribution, even if some of the involved objects or skills are familiar. Large deviations in object placement, relative spatial layout, or scene geometry can make the generated action sequence less reliable and may cause errors to accumulate over replanning steps. These observations suggest that the compositional generalization of W2A is still bounded by the support of the training distribution. Since W2A distills world priors into the action model and removes the video model at test time, it does not retain explicit online future imagination or strong recovery mechanisms for out-of-distribution situations. As a result, its robustness remains limited when both semantic novelty and geometric distribution shift become significant.

\section{Discussion of Related Efforts on Improving WAM Efficiency}
\label{app:discussion_of_related_efforts_on_improving_wam_efficiency}
Several recent video-action and world-action models have explored how to reduce the inference cost introduced by test-time world modeling. For example, UVA~\citep{li2025unified} improves action inference efficiency by learning a joint video-action latent representation while decoupling video and action decoding, allowing action prediction to bypass explicit future video generation at inference time. Other efficient WAM designs reduce deployment cost by reducing or skipping explicit future prediction~\citep{yuan2026fastwam}, or by adopting an action-centered formulation in which future video generation becomes optional rather than a prerequisite for action decoding~\citep{ye2026gigaworld}. These works suggest that the benefits of video-based world modeling may come not only from test-time imagination, but also from the world-aware representations learned during training.

W2A is complementary to these efforts, but differs in its deployment-time computation graph. Rather than designing a more efficient world-action inference pipeline, W2A uses the video DiT only as a training-time teacher and distills its intermediate denoising representations into an action DiT. After training, the deployed policy does not run video generation, predict future visual latents, or execute a separate world-modeling branch. This makes W2A closer to a standard action model at deployment, while still benefiting from video-based world priors during training.

In summary, existing efficient WAMs mainly ask how to make test-time world modeling cheaper, whereas W2A asks whether useful world priors can be internalized into the action model so that test-time world modeling is no longer required.

\end{document}